\documentclass[twoside,11pt]{article}

\usepackage{jmlr2e}

\usepackage{graphicx}
\usepackage{xcolor}
\usepackage{amstext}
\usepackage{amsmath}
\usepackage{amssymb}
\usepackage{booktabs}
\usepackage{subfigure}
\newtheorem{myDef}{Definition}

\newtheorem{prop}{Proposition}

\usepackage[flushleft]{threeparttable}
\usepackage{array, multirow, multicol}

\usepackage{algorithm}
\usepackage{algpseudocode}

\algnewcommand\algorithmicinput{\textbf{Input:}}
\algnewcommand\Input{\item[\algorithmicinput]}
\algnewcommand\algorithmicoutput{\textbf{Output:}}
\algnewcommand\Output{\item[\algorithmicoutput]}

\jmlrheading{X}{XXXX}{1-XX}{XX/XX}{XX/XX}{XXX}{Kuo Gai and Shihua Zhang}

\ShortHeadings{Deep Residual Networks Learn the Geodesic Curve in the Wasserstein Space}{Kuo Gai and Shihua Zhang}
\firstpageno{1}

\begin{document}
		
		\title{Deep Residual Networks Learn the Geodesic Curve in the Wasserstein Space}
		
		\author{\name Kuo Gai$^{1}$ and %\email gaikuo@amss.ac.cn; 
			\name Shihua Zhang$^{2,3,*}$ \\ %\email zsh@amss.ac.cn   \\
			\addr 
            $^{1}$Shanghai Institute for Mathematics and Interdisciplinary Sciences (SIMIS)\\
			Shanghai 200433, China\\ 
            $^{2}$Academy of Mathematics and Systems Science\\
			Chinese Academy of Sciences\\
			Beijing 100190, China\\
			$^{3}$School of Mathematical Sciences\\
			University of Chinese Academy of Sciences\\
			Beijing 100049, China
%			\AND
%			\name Shihua Zhang \email zsh@amss.ac.cn \\
%			\addr Academy of Mathematics and Systems Science\\
%			Chinese Academy of Sciences\\
%			Beijing 100190, China\\
%			School of Mathematical Sciences\\
%			University of Chinese Academy of Sciences\\
%			Beijing 100049, China
%           \\ Correspondence: \email zsh@amss.ac.cn
            \\ $^*$Email: zsh@amss.ac.cn
		}
		
	%	\editor{XXX}
		
		\maketitle

\begin{abstract}
Recent studies revealed the mathematical connection between deep neural networks (DNNs) and dynamic systems. However, the specific dynamics that DNNs, especially deep residual networks (ResNets), tend to learn during training remain insufficiently characterized. To this end, we model the forward propagation of deep residual networks using continuity equations, in which the measure is conserved and infinite curves in the measure space connect the input distribution to the output one of a ResNet. We find ResNets with $L_2$ regularization attempt to learn the geodesic curve in the Wasserstein space, induced by the optimal transport map. Compared with plain networks, ResNets can better approximate the geodesic curve, which explains why ResNets can be optimized and generalize better. Numerical experiments show that the data tracks of a ResNet tend to be line-shaped in terms of the line-shape score, and the map learned by a ResNet is closer to the optimal transport map in terms of the optimal transport score. In a word, we conclude that ResNets learn the geodesic curve in the Wasserstein space and discretely engineer the data transformation in high-dimensional spaces.  
\end{abstract}
		
% Note that keywords are not normally used for peerreview papers.

\clearpage
\section{Introduction}\label{sec:introduction}

Deep neural networks (a.k.a. deep learning) have become a core modeling paradigm in computer vision, natural language processing, computational biology, and, more recently, large-scale foundation models. Despite these remarkable successes, modern DNNs remain difficult to analyze and verify, especially in safety- and security-sensitive scenarios, because their highly compositional architectures combine nonlinear activations and massive parameterization, leading to complex non-convex optimization landscapes and opaque internal representations. A growing body of theoretical work has sought to understand DNNs through idealized or limiting models, including neural tangent kernels \citep{jacot2018neural}\citep{arora2019fine}\citep{du2018gradient}\citep{du2019gradient}, deep linear networks \citep{saxe2013exact}\citep{arora2019convergence}\citep{du2019width}, and generalizing existing machine learning techniques such as matrix decomposition and sparse coding to multi-layer ones \citep{arora2019implicit}\citep{papyan2017convolutional}. These models provide valuable insights, but they often rely on simplifications that do not fully characterize the behavior of practical deep architectures.

\par Residual connections remain central to this development. They were originally crucial for training very deep convolutional networks such as ResNets \citep{he2016deep}. But they are also a fundamental component of the Transformer architecture: each self-attention or feed-forward sublayer is wrapped by a residual connection and normalization \citep{vaswani2017attention}. This observation is especially important because most modern LLMs can be viewed as very deep stacks of residual Transformer blocks. Subsequent studies further show that normalization placement, residual-branch design, and extensions such as Manifold-Constrained Hyper-Connections are crucial for stabilizing and scaling deep Transformers and foundation models beyond vision-based ResNets \citep{xiong2020layer,xie2026mhc}. Therefore, understanding the dynamics learned by residual architectures is no longer only a question about ResNets in computer vision but also about the backbone of modern foundation models. Although ResNets have been interpreted from several perspectives, including the unraveled view \citep{veit2016residual}, the unrolled iterative estimation view \citep{greff2016highway}\citep{jastrzebski2018residual}, the multi-layer convolutional sparse coding view \citep{zhang2019towards}, and the dynamical-system view \citep{weinan2017proposal}\citep{haber2017stable}, a theoretical insight into why residual architectures outperform plain networks, and what dynamics they tend to learn during training, remains open.

\par In computer science, continuous concepts are usually realized through discretization. In contrast, many discrete phenomena are analyzed in continuous manners. Following this principle, to analyze ResNets theoretically, one can assume: (1) the data points are sampled from a continuous distribution; (2) the layerwise transformation of a ResNets can be viewed as a discrete approximation to the continuous curve in the probability measure space $\mathcal{P}(\mathbb{R}^d)$, which can be described with a dynamic system. E \citep{weinan2017proposal} and Haber and Ruthotto \citep{haber2017stable} first explained ResNets with ordinary differential equations (ODEs). Wang et al. \citep{wang2019resnets} showed a connection between ResNets and transport equations (TEs) to investigate data flow in both forward and backward propagations. However, because the proportions of data points in certain classes are fixed across different layers' representation spaces, both ODE and TE fail to model the conservation of probability during the forward propagation of ResNets.

\begin{figure}[t!]
	\centering
	\includegraphics[width=1.0\columnwidth]{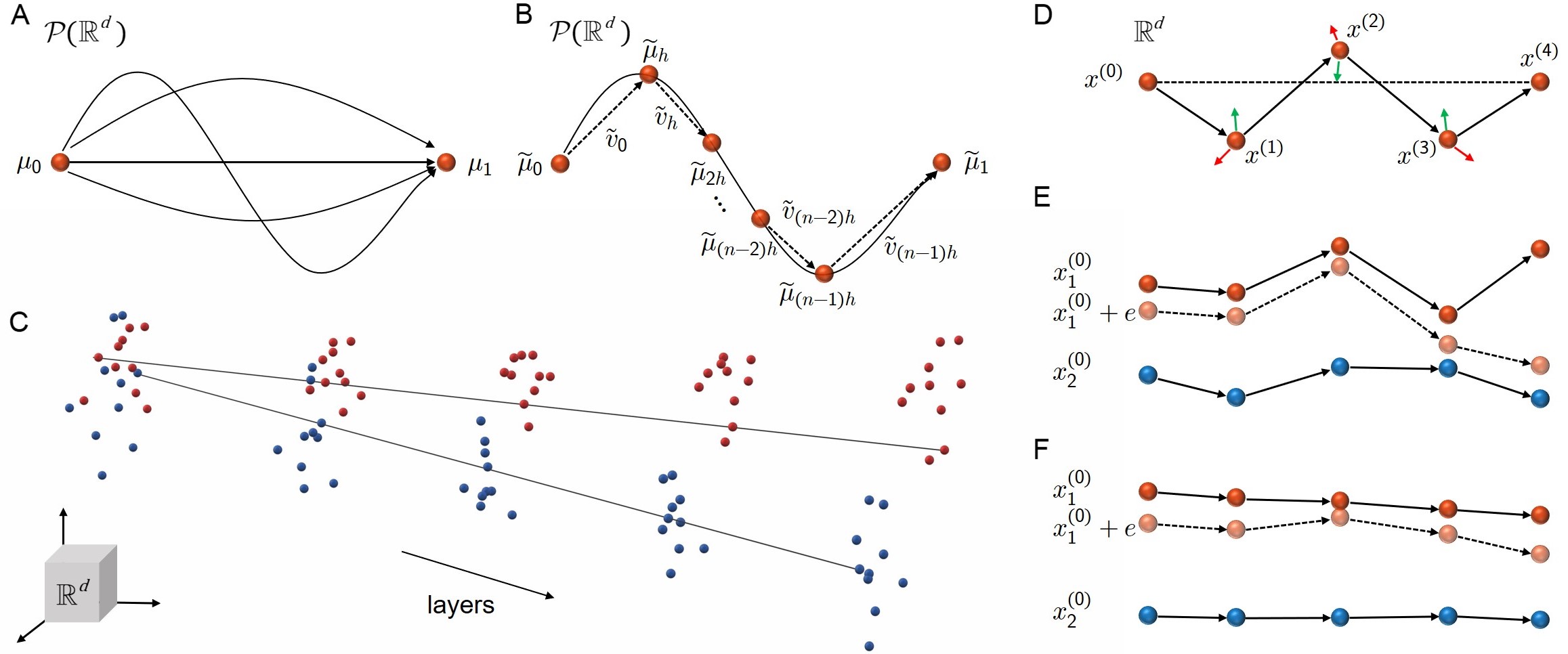}
	\caption{Illustration of learning the geodesic curve in the Wasserstein space with a ResNet. (A) In the measure space $\mathcal{P}(\mathbb{R}^d)$, there are infinite curves connecting $\mu_0$ and $\mu_1$ which satisfy continuity equations in (\ref{eq:continiuty equation}). (B) A ResNet is a discretization of a continuous curve. The vector fields $\tilde{v}_0,\tilde{v}_h,\cdots,\tilde{v}_{(n-1)h}$ are learned by shallow networks, where $h=\frac{1}{n}$. Then we obtain a polyline connecting $\tilde{\mu}_0,\tilde{\mu}_h,\cdots,\tilde{\mu}_{(n-1)h},\tilde{\mu}_1$. (C) The data points are forward propagated through a straight line inside a ResNet of four layers according to the geodesic curve induced by the optimal transport map.  
(D) The red arrows indicate that the data tracks tend to deviate from the straight line from the data $x^{(0)}$ to its representation $x^{(4)}$. The green arrows indicate that $L_2$ regularization (or weight decay) in a ResNet can draw the tracks to the straight line. 
(E) Two data tracks intersect or close, then the two data points tend to be indistinguishable when adding noise. (F) Two data points, even with noise, can be well discriminated when the data tracks are organized with an optimal transport map.  }
	\label{fig1}
\end{figure}

\begin{figure}[t!]
	\centering
	\includegraphics[width=0.94\columnwidth]{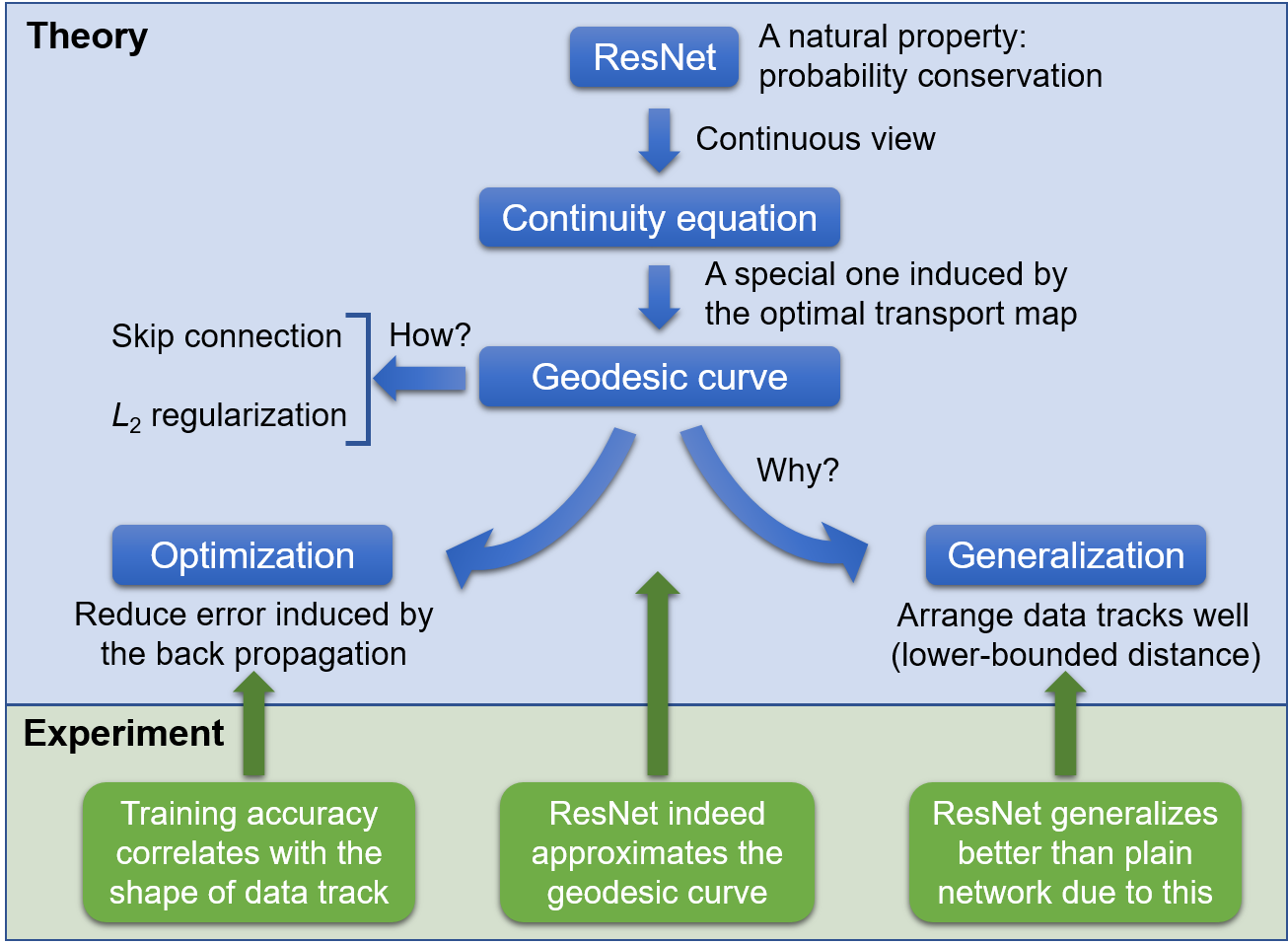}
	\caption{The roadmap of this paper. }
	\label{fig17}
\end{figure}

\par To address this issue, we utilized the curves satisfying continuity equations which conserve the measure of distributions to model ResNets (Fig. 1A). Among all the curves, the geodesic one in the Wasserstein space $(\mathcal{P}_2(\mathbb{R}^d),W_2)$ has very good properties which can benefit the optimization and generalization of ResNets. A ResNet can be considered as a layerwise transformation to approximate the continuous transformation from $\mu_{0}$ and $\mu_1$ which denote the continuous distributions corresponding to the data and its representations respectively (Fig. 1B). According to the Benamou-Brenier formula \citep{benamou1999numerical}, the approximation of a ResNet and the geodesic curve can be achieved by minimizing an energy function which is upper bounded by the $L_2$ regularization. Thus, a ResNet has two characteristics (Fig. 1C), i.e., (1) the data tracks of layerwise transformation of a ResNet tend to be line-shaped; (2) the map learned by a ResNet is close to the optimal transport map.

\par Compared with plain networks, ResNets can better approximate the geodesic curve according to the forward Euler method. It provides a possible geometric mechanism for why residual architectures are easier to optimize and often exhibit better generalization and robustness. First, the data tracks in ResNets tend to be more line-shaped than those in plain networks, though both likely zigzag due to the back-propagation (BP) algorithm (Fig. 1D). We demonstrate that the optimization process is more effective along a straight line than a zigzag track. Thus, ResNets can overcome the degradation problem and achieve lower loss by employing more layers. Second, since the maps learned by ResNets are closer to the optimal transport map, the distance between two tracks is lower-bounded. Therefore, ResNets do not mix up two tracks when adding noise (Fig. 1E and F) and show better generalization. 

\par Numerical experiments show that the data tracks of a ResNet indeed tend to be line-shaped in terms of line-shape score (LSS), and the map learned by a ResNet is closer to the optimal transport map in terms of optimal transport score (OTS). Moreover, LSS and OTS are closely related to the performance of ResNets, suggesting their principle is indeed to learn the geodesic curve in the Wasserstein space. For convenience, we illustrate the roadmap of this paper (Fig. \ref{fig17}).

\section{Methodology}
In this section, we first assume the output dimension of a ResNet is the same as the input dimension to derive the connection of a ResNet and the geodesic curve in the Wasserstein space, and then discuss the generalization to dimension-varying situations in subsection 2.3.

\subsection{Connection of ResNets and the continuity equations}

The connection between ResNets and dynamic systems is first explained by Haber and Ruthotto \citep{haber2017stable} and E \citep{weinan2017proposal}. The general form of a ResNet block in forward propagation is
\begin{equation}
x^{(k+1)}=x^{(k)}+v_k(x^{(k)}),\  k\in \{0,1,\cdots,n\},
\end{equation}
where $v_k(\cdot)$ is a function induced by a shallow network, $x^{(k+1)}$ and $x^{(k)}$ are the outputs of the $k$th and $(k+1)$th ResNet blocks, respectively. It can be viewed as a discretization of the dynamical system
\begin{equation}
\label{eq:dynamic}
\frac{dx}{dt}=v(x,t),\ t\in[0,1].
\end{equation}
Suppose we have a dataset $\mathcal{D}=\{x_1,\cdots,x_m\}$, a general assumption for a ResNet says that when $m$ is large, for each $k$, the $k$th level of representation of data points $x_i^{(k)}$, $i\in \{1,\cdots,m\}$ are sampled from continuous distributions, denoted by $\mu_k$. Then the corresponding dynamic system has the form 
\begin{equation}
\label{eq:dynamic_distribution}
\frac{d\mu_t}{dt}=\tilde{v}(\mu_t,t),\ t\in [0,1].
\end{equation}
\par Next, we note that the dynamic system in (\ref{eq:dynamic_distribution}) should conserve mass. For each subset $S\subset\mathcal{D}$ and a fixed $k$, the empirical probability of the $k$th level representation $x_i^{(k)}$ belonging to the set $\{x_j^{(k)}|x_j\in S\}$ is
\begin{equation}
P\left( \{x_j^{(k)}|x_j\in S\}\right)=\frac{|\{x_j^{(k)}|x_j\in S\}|}{|\{x_j^{(k)}|x_j\in \mathcal{D}\}|}=\frac{|S|}{|\mathcal{D}|},
\end{equation}  
which does not depend on $k$. It is a natural property of ResNets (and general DNNs) and should be preserved in the setting of continuous distributions and dynamic systems with continuous time.
\par Thus, the dynamic system we use to model a ResNet should conserve probability. For instance, if $R\subset \operatorname{supp}(\mu_0)$ and $R_t$ denotes the set mapped from $R$ by the dynamic system (\ref{eq:dynamic_distribution}), then
\begin{equation}
\mu_t(R_t)=\mu_{0}(R),
\end{equation}
which does not depend on $t$. A dynamic system that conserves mass is termed the continuity equation (Fig. 1A), which has the form
\begin{equation}\label{eq:continiuty equation}
\frac{d}{dt}\mu_t+\nabla\cdot(v_t\mu_t)=0,
\end{equation}
where the vector field $v_t$ is the infinitesimal variation of the continuum $\mu_t$. Its   discrete form is   
\begin{equation}
\tilde{\mu}_{(i+1)h}=(I_d+\tilde{v}_{i})_{\#}\tilde{\mu}_{ih},\quad i =0,1,\cdots, n-1,
\end{equation}
where $I_d$ denotes the identity map, $\tilde{\mu}_{ih} \ (i=0,1,\cdots,n)$ are discrete points on the curve $(\mu_t)$ (Fig. 1B). In a plain network, $I_d+\tilde{v}_i$ corresponds to a single layer map. Compared with the plain network, ResNet is a better approximation to the continuity equation (\ref{eq:continiuty equation}) because $\tilde{v}_i$ is induced by a shallow network and $I_d$ is modeled by a skip connection (Fig. 2).

\subsection{Geodesic curve in the Wasserstein space}
When $\mu_0$ and $\mu_1$ are fixed, there are infinite curves in the probability measure space connecting $\mu_0$ and $\mu_1$ (Fig. 1A), each of which can be approximated by a ResNet (Fig. 1B). However, not all the networks can be optimized or generalize well (see Sections 3 and 4 in detail). Among all the curves, the geodesic one in the Wasserstein space has very good properties. It is induced by the optimal transport map $T$ from a distribution $\mu_0$ to another distribution $\mu_1$ with the minimum cost. This cost is termed as the Wasserstein distance $W(\mu_0,\mu_1)$, i.e.
\begin{equation}
W_c(\mu_{0},\mu_1)=\inf_{T_{\#}\mu_{0}=\mu_1}\int c\left(x,T\left(x\right)\right)d \mu_0\left(x\right),
\end{equation}
where $c(\cdot,\cdot)$ is the cost function of moving each unit mass from the source $\mu_0$ to the target $\mu_1$. In this paper, $c\left(x,T\left(x\right)\right)$ is set to $\|x-T(x)\|_2^2$, which is commonly used in both optimal transport and deep learning. The geodesic curve ($\mu_t$) is induced by the optimal transport map $T$, i.e. 
\begin{equation}
\label{eq: optimal curve}
\mu_t=\left(\left(1-t\right)I_d+tT\right)_{\#}\mu_0.
\end{equation}
For each point $x\in \operatorname{supp}(\mu_0)$, $x$ is transported through a straight line $x^{(t)}=(1-t)x+tT(x)$. Since ($\mu_t$) satisfies the continuity equation (\ref{eq:continiuty equation}), the construction of ($\mu_{t}$) in (\ref{eq: optimal curve}) results in the specific structure of the vector field $(v_t)$, i.e.,
\begin{equation}
\label{eq: optimal vector field}
v_{t}:=(T-I_d) \circ((1-t) I_d+t T)^{-1},
\end{equation}
However, it is not easy to impose the constraint induced by (\ref{eq: optimal vector field}) on $\tilde{v}_{i}$ directly, because in the training process of ResNets, we actually don't know $T$. Also, it is intractable to force $\tilde{v}_{i}$ to satisfy (\ref{eq: optimal vector field}) because it has too many parameters. Alternatively, the following proposition transforms this constraint into an energy-function minimization problem. For the detailed arguments on optimal transport, we refer readers to Chapter 2 of \cite{ambrosio2013user}.

\begin{prop}
	(\textbf{Benamou-Brenier formula\citep{benamou1999numerical}})
Let $\mu_0,\mu_1\in\mathcal P_2(\mathbb R^d)$. Then
\begin{equation}
	\label{eq:bbformula}
	W_2^2(\mu_0,\mu_1)
	=
	\inf_{(\mu_t,v_t)}
	\int_0^1
	\|v_t\|_{L^2(\mu_t)}^2dt,
\end{equation}
where the infimum is taken over all weakly continuous curves $(\mu_t)_{t\in[0,1]}$
and velocity fields $(v_t)_{t\in[0,1]}$ satisfying the continuity equation
\[
\partial_t\mu_t+\nabla\cdot(v_t\mu_t)=0,
\qquad
\mu_{t=0}=\mu_0,\quad \mu_{t=1}=\mu_1.
\]
\end{prop}

Let $h=1/n$. A ResNet block has the form
\[
x^{(i+1)}=x^{(i)}+\tilde v_i(x^{(i)}),
\]
which can be viewed as a forward Euler discretization of the continuous flow with
$\tilde v_i\approx h v_{ih}$. Therefore, the discrete counterpart of the Benamou--Brenier
energy is
\begin{equation}
	\label{pb:optimal}
	\begin{aligned}
		&\min_{\tilde v_i}\quad
		\frac{1}{h}\sum_{i=0}^{n-1}
		\|\tilde v_i\|_{L^2(\tilde\mu_{ih})}^2\\
		&\mbox{s.t.}\quad
		\left\{
		\begin{aligned}
			&\tilde\mu_0=\mu_0,\\
			&\tilde\mu_{(i+1)h}=(I_d+\tilde v_i)_\#\tilde\mu_{ih},
			\quad i=0,1,\cdots,n-1,\\
			&\tilde\mu_1=\mu_1.
		\end{aligned}
		\right.
	\end{aligned}
\end{equation}

For a ResNet, let
\[
\tilde v_i(\cdot)=W_i^{(2)}\sigma(W_i^{(1)}\sigma(\cdot)),
\]
where $W_i^{(1)},W_i^{(2)}$ are weight matrices and $\sigma(\cdot)$ is the ReLU activation.
Let $\|\cdot\|_{\mathrm{op}}$ denote the spectral norm and $\|\cdot\|_F$ denote the
Frobenius norm. Since ReLU is non-expansive and
$\|W\|_{\mathrm{op}}\leq \|W\|_F$, we have
\begin{equation}
	\label{eq:wdbound_res}
	\begin{aligned}
		\|\tilde v_i\|_{L^2(\tilde\mu_{ih})}^2
		&=
		\int
		\|W_i^{(2)}\sigma(W_i^{(1)}\sigma(x))\|_2^2
		d\tilde\mu_{ih}(x)\\
		&\leq
		\|W_i^{(2)}\|_{\mathrm{op}}^2
		\|W_i^{(1)}\|_{\mathrm{op}}^2
		\int\|x\|_2^2d\tilde\mu_{ih}(x)\\
		&\leq
		\|W_i^{(2)}\|_F^2
		\|W_i^{(1)}\|_F^2
		\int\|x\|_2^2d\tilde\mu_{ih}(x)\\
		&\leq
		\frac{1}{4}
		\left(
		\|W_i^{(2)}\|_F^2+\|W_i^{(1)}\|_F^2
		\right)^2
		\int\|x\|_2^2d\tilde\mu_{ih}(x).
	\end{aligned}
\end{equation}

Assume that during training there exist constants $M>0$ and $R>0$ such that
\[
\int\|x\|_2^2d\tilde\mu_{ih}(x)\leq M
\]
and
\[
\|W_i^{(1)}\|_F^2+\|W_i^{(2)}\|_F^2\leq R
\]
for all $i$. Then, by (\ref{eq:wdbound_res}), there exists a constant $C>0$ such that
\begin{equation}
	\label{eq:weight}
	\frac{1}{h}
	\sum_{i=0}^{n-1}
	\|\tilde v_i\|_{L^2(\tilde\mu_{ih})}^2
	\leq
	\frac{C}{h}
	\sum_{i=0}^{n-1}
	\left(
	\|W_i^{(1)}\|_F^2+\|W_i^{(2)}\|_F^2
	\right).
\end{equation}

For a fixed depth $n$, the factor $1/h$ is a constant. Hence, the right-hand side has the
same form as the weight decay penalty. It suggests that weight decay biases ResNets
toward low-energy transport paths in the sense of the discretized Benamou--Brenier action.

For a plain network, let
\[
(I_d+\tilde v_i)(\cdot)=\sigma(W^{(i)}\cdot).
\]
Then
\[
\tilde v_i(x)=\sigma(W^{(i)}x)-x.
\]
Different from ResNets, even when $W^{(i)}=0$, the induced displacement is
$\tilde v_i(x)=-x$, which is not controlled by the weight decay penalty. Indeed,
\begin{equation}
	\label{eq:plain_weight_decay}
	\begin{aligned}
		\|\tilde v_i\|_{L^2(\tilde\mu_{ih})}^2
		&=
		\int
		\|\sigma(W^{(i)}x)-x\|_2^2d\tilde\mu_{ih}(x)\\
		&\leq
		2\int
		\left(
		\|\sigma(W^{(i)}x)\|_2^2+\|x\|_2^2
		\right)d\tilde\mu_{ih}(x)\\
		&\leq
		2\left(\|W^{(i)}\|_{\mathrm{op}}^2+1\right)
		\int\|x\|_2^2d\tilde\mu_{ih}(x)\\
		&\leq
		2\left(\|W^{(i)}\|_F^2+1\right)
		\int\|x\|_2^2d\tilde\mu_{ih}(x).
	\end{aligned}
\end{equation}

Assume that
\[
\int\|x\|_2^2d\tilde\mu_{ih}(x)\leq M
\]
for all $i$. Then
\begin{equation}
	\label{eq:plain_weight_decay_sum}
	\frac{1}{h}
	\sum_{i=0}^{n-1}
	\|\tilde v_i\|_{L^2(\tilde\mu_{ih})}^2
	\leq
	\frac{C'}{h}
	\sum_{i=0}^{n-1}
	\left(1+\|W^{(i)}\|_F^2\right).
\end{equation}

The additional constant term is absent in the residual case. Therefore, weight decay in a plain network does not control the discretized transport energy as directly as in a ResNet. This reflects the fact that a plain layer has to learn both the identity component and the residual displacement, while a ResNet provides the identity component explicitly through the skip connection.

We denote $:\mathbb{R}^d\to\mathbb{R}^d$ as the map learned by stacking $n$ ResNet blocks, i.e. \begin{equation}
\label{eq:ResNet}
f=(I_d+\tilde{v}_{n-1})\circ(I_d+\tilde{v}_{n-2})\cdots\circ(I_d+\tilde{v}_{0}).
\end{equation}
Let $\mathcal{L}(f)$ be the loss function of a certain task with respect to $f$. Then the problem of solving the task with a ResNet can be formulated as
\begin{equation}
\label{eq:res optimize}
\min_{f}\mathcal{L}(f)+\gamma \sum_{W\in \operatorname{Para}   \{f\}   }\|W\|_F^2,
\end{equation} 
where $\gamma$ is the hyperparameter to balance the two terms in the objective function and $\operatorname{Para}\{f\}$ denotes the set of weight matrices in $f$. By adding the regularizer, if the objective in (\ref{eq:res optimize}) is well-optimized, then the model can simultaneously achieve: (1) a map which can solve the task with low loss; (2) a discrete approximation to a curve from the input distribution to the output distribution of $f$ which is close to the geodesic curve induced by the optimal transport map. Numerical experiments in Section 5 show that ResNets with weight decay indeed approximate a geodesic curve.

\subsection{Generalization to dimension-varying situations}
\par The connection of ResNets and geodesic curve in the Wasserstein space is limited to the case where the dimensions of input and output distributions are the same. However, in general, the number of labels is much smaller than the data dimension, which means most of the information in the data is redundant. A natural solution is to reduce the layer dimensions gradually to the label number to exclude redundant information. Let $f_k: \mathbb{R}^{d_k}\to \mathbb{R}^{d_k}$ $ (k=1,2,\cdots,K)$ denote the function composed of multiple layers or blocks. Let $P_k: \mathbb{R}^{d_k}\to \mathbb{R}^{d_{k+1}}$ $(k=1,2,\cdots,K-1)$ denote the function to change the dimension from $d_k$ to $d_{k+1}$ such as convolutions and poolings. For instance, in ResNet, $P_k$ is constructed by replacing the identity map in the ResNet block with a convolution operation with kernel size equaling 1. Formally, let  
\begin{equation}
\label{eq:dimension_reduction}
F=f_K\circ P_{K-1}\circ f_{K-1}\cdots P_1\circ f_1,
\end{equation}
the optimization problem of ResNet with dimension reduction of $K-1$ times is 
\begin{equation}
\label{pb:multi_dimension}
\min_{F}\mathcal{L}(F)+\gamma \sum_{W\in \operatorname{Para}\{F\}}\|W\|_F^2,
\end{equation}
where $\operatorname{Para}\{F\}$ denotes the set of weight matrices in $F$.

\section{Optimization}
\par In a ResNet, each data point follows a layerwise trajectory induced by the sequence of transformations. We refer to this trajectory as the data track. Empirically, we observe that the optimization performance improves when data tracks become closer to straight lines (Section 5.3). This observation motivates the hypothesis that networks with straighter data tracks are easier to optimize, because their layerwise transformations move data points in more coherent directions. It is a surprising argument since, for traditional machine learning models such as linear regression, adding a regularizer will increase the training loss.

\begin{myDef}
	(\textbf{Data track}) For $f$ in (\ref{eq:ResNet}), let $x^{(0)}=x$ and
	\[
	x^{(l+1)}=(I_d+\tilde{v}_l)(x^{(l)}), \quad l=0,1,\cdots,n-1.
	\]
	The data track of $x$ corresponding to $f$ is defined as the sequence
	\[
	x^{(0)}\to x^{(1)}\to\cdots\to x^{(n)},
	\]
	denoted by $\operatorname{track}_{f}(x)$.
\end{myDef}

To quantify how close a data track is to a straight line, we define the line-shape score in (\ref{score:lss}) and line shape ratio in (\ref{score:lsr}). A value close to one indicates that the data track is nearly straight, while a larger value suggests a more curved or oscillatory trajectory.

A possible explanation can be obtained from a local linearization of the loss. Let
\[
g(x)=\nabla_{x^{(n)}}\ell(x^{(n)},y)
\]
be the gradient of the loss with respect to the final representation, and let
\[
u_g(x)=-\frac{g(x)}{\|g(x)\|_2}
\]
be the steepest descent direction in the output space. Suppose that a small update of the
$l$th residual block induces an effective perturbation $q_l(x)$ on the final representation.
Then, under the first-order approximation,
\[
\ell\left(x^{(n)}+\sum_{l=0}^{n-1}q_l(x),y\right)
-
\ell(x^{(n)},y)
=
\left\langle g(x),\sum_{l=0}^{n-1}q_l(x)\right\rangle
+
O\left(\left\|\sum_{l=0}^{n-1}q_l(x)\right\|_2^2\right).
\]
Decompose each effective perturbation into the component parallel to $u_g(x)$ and the
orthogonal component:
\[
q_l(x)=a_l(x)u_g(x)+b_l(x),
\qquad b_l(x)\perp u_g(x).
\]
Then the first-order decrease of the loss is
\[
\left\langle g(x),\sum_{l=0}^{n-1}q_l(x)\right\rangle
=
-\|g(x)\|_2\sum_{l=0}^{n-1}a_l(x).
\]
Therefore, only the components parallel to the descent direction contribute to the first-order decrease of the loss. While the orthogonal components consume update energy, they do not directly reduce the loss.

Assume the total size of the effective layerwise perturbations is bounded by
\[
B(x)=\sum_{l=0}^{n-1}\|q_l(x)\|_2^2.
\]
Since
\[
B(x)=\sum_{l=0}^{n-1}a_l(x)^2+\sum_{l=0}^{n-1}\|b_l(x)\|_2^2,
\]
we have, by Cauchy's inequality,
\[
\left(\sum_{l=0}^{n-1}a_l(x)\right)^2
\leq
n\sum_{l=0}^{n-1}a_l(x)^2
=
n\left(B(x)-\sum_{l=0}^{n-1}\|b_l(x)\|_2^2\right).
\]
Thus, for a fixed update budget $B(x)$, the orthogonal energy
$\sum_l\|b_l(x)\|_2^2$ reduces the upper bound of the achievable first-order
decrease. The most efficient case corresponds to
\[
q_0(x)=q_1(x)=\cdots=q_{n-1}(x)
=
\sqrt{\frac{B(x)}{n}}\,u_g(x),
\]
or equivalently,
\[
b_l(x)=0,\qquad
a_l(x)=\sqrt{\frac{B(x)}{n}},\qquad l=0,\ldots,n-1.
\] In this case, the update budget is concentrated in the one-dimensional subspace spanned by $u_g(x)$ and is used most efficiently for reducing the loss.

\section{Generalization}
\par High-dimensional data is generally assumed to be concentrated on a low-dimensional manifold embedded in the high-dimensional background space. DNN should not only fit the training data, but also generalize to the untrained region surrounding them. In this section, we illustrate how the generalization benefits from the principle of approximating the geodesic curve.  
\par According to the study in \citep{xu2012robustness}, the generalization error has an upper bound defined by the robustness of a learning algorithm. For fixed data point $x_i$ with data track $x_i^{(0)}\to x_i^{(1)} \cdots \to x_i^{(n)}$, after adding a noise $\epsilon$ to $x_i$, the data track is changed to $x_i^{(0)}+\epsilon^{(0)}\to x_i^{(1)}+\epsilon^{(1)} \cdots \to x_i^{(n)}+\epsilon^{(n)}$, where $\epsilon^{(l)}$ denote the variation of $l$th representation $x_i^{(l)}$. Besides $x_i^{(0)}\to x_i^{(1)} \cdots \to x_i^{(n)}$, there are $m$ data tracks in $\mathbb{R}^d$. They should be arranged far from each other to avoid being mixed up by noise. We first define the distance between two data tracks for a learned map $f$ of a ResNet. 
\begin{myDef}
	(\textbf{distance of tracks}) The distance of two data tracks $\operatorname{track}_{f}(x_i)$ and $\operatorname{track}_{f}(x_j)$ is defined as the minimum $L_2$ distance of data points in $\operatorname{track}_{f}(x_i)$ and $\operatorname{track}_{f}(x_j)$ with the same index, i.e.,
	\begin{equation}
	d(\operatorname{track}_{f}(x_i),\operatorname{track}_{f}(x_j))=\min_{l=0}^n \|x_i^{(l)}-x_j^{(l)}\|_2.
	\end{equation}
\end{myDef}
In the following, we derive that, if the ResNet approximates the geodesic curve, for arbitrary $i,j\in \{1,2,\cdots,m\}$, the distance of two tracks $\operatorname{track}_{f}(x_i),\operatorname{track}_{f}(x_j)$ is lower bounded. 
\begin{theorem}
	Suppose the map $f$ of a ResNet is the optimal transport map from the input distribution to output one, $\forall i,j\in \{1,2,\cdots,m\}$, let $x_i^{(t)}=\left(1-t\right)x_i+tf(x_i),x_j^{(t)}=\left(1-t\right)x_j+tf(x_j)$. Then we have 
	\begin{equation}
	\operatorname{min}_t\|x_i^{(t)}-x_j^{(t)}\|_2\geq \frac{\|x_i-x_j\|_2\|f(x_i)-f(x_j)\|_2}{\sqrt{\|x_i-x_j\|_2^2+\|f(x_i)-f(x_j)\|_2^2}}.
	\end{equation}
	which means 
	\begin{equation}
	\label{eq:lower_bound}
	d(\operatorname{track}_{f}(x_i),\operatorname{track}_{f}(x_j))\geq \frac{\|x_i-x_j\|_2\|f(x_i)-f(x_j)\|_2}{\sqrt{\|x_i-x_j\|_2^2+\|f(x_i)-f(x_j)\|_2^2}}.
	\end{equation}
\end{theorem}    

\begin{proof}
	Since $f$ is the optimal transport map with the minimum cost, it holds
	\begin{equation}
	\label{eq:minimum cost}
	\|f(x_i)-x_i\|^2_2+\|f(x_j)-x_j\|^2_2\leq \|f(x_i)-x_j\|^2_2+\|f(x_j)-x_i\|^2_2.
	\end{equation}
	After simplifying the equality (\ref{eq:minimum cost}), we have 
	\begin{equation}
	<f(x_i)-f(x_j),x_i-x_j>\geq 0.
	\end{equation}
	Observe that
	\begin{equation}
	\label{eq:lower bound}
	\begin{aligned}
	\|x_i^{(t)}-x_j^{(t)}\|_2^2&=\|(1-t)x_i+tf(x_i)-(1-t)x_j-tf(x_j)\|_2^2\\
	&=(1-t)^2\|x_i-x_j\|_2^2+t^2\|f(x_i)-f(x_j)\|_2^2\\
	&+2t(1-t)<x_i-x_j,f(x_i)-f(x_j)>\\
	&\geq (1-t)^2\|x_i-x_j\|_2^2+t^2\|f(x_i)-f(x_j)\|_2^2
	\end{aligned},
	\end{equation}
	when $t=\frac{\|x_i-x_j\|_2^2}{\|x_i-x_j\|_2^2+\|f(x_i)-f(x_j)\|_2^2}$, the right side of (\ref{eq:lower bound}) takes the minimum. Thus we have
	\begin{equation}
	\begin{aligned}
	\operatorname{min}_t\|x_i^{(t)}-x_j^{(t)}\|_2\geq \frac{\|x_i-x_j\|_2\|f(x_i)-f(x_j)\|_2}{\sqrt{\|x_i-x_j\|^2_2+\|f(x_i)-f(x_j)\|^2_2}}
	\end{aligned}.
	\end{equation}
\end{proof}
When $x_i$ and $x_j$ belong to different classes, both $\|x_i-x_j\|_2$ and $\|f(x_i)-f(x_j)\|_2$ should be relatively large. According to (\ref{eq:lower_bound}), $d(\operatorname{track}_{f}(x_i),\operatorname{track}_{f}(x_j))$ is lower bounded. Thus, $f$ does not easily mix up $x_i$ and $x_j$.
\par As we have discussed before, ResNet is a better approximation to the geodesic curve in the Wasserstein space. It provides a possible geometric mechanism for the improved robustness and generalization of residual architectures, consistent with our numerical experiments in Section 5.

\section{Results}
\label{sec:results}
\subsection{Experimental setup}
In the curve induced by the optimal transport map $T$, each point $x\in\mu_0$ is transported along a straight line $x^{(t)}=(1-t)x+tTx$. Thus, to numerically evaluate the principle that ResNets indeed approximately learn the geodesic curve in the Wasserstein space, we need to test: 1) whether the data tracks in ResNets are indeed close to the straight lines or not; and 2) whether the map learned by ResNets is close to the optimal transport map or not. 

\par \textbf{Line-shape score (LSS)}
First, we define the line-shape ratio (LSR) and line-shape score (LSS) to measure the closeness of the data tracks in the plain network and ResNet to straight lines. Specifically, 
\begin{equation}\label{score:lsr}
	\operatorname{LSR}=\frac{\sum_{l=0}^{n-1}\|x^{(l+1)}-x^{(l)}\|_2}{\|x^{(n)}-x^{(0)}\|_2}.
\end{equation}
However, if one segment takes up a great proportion of the track length, i.e., for some $j$, $\|x^{(j+1)}-x^{(j)}\|_2$ is much larger than $\|x^{(i+1)}-x^{(i)}\|_2\ (i\neq j)$, then LSR can still be close to 1. For instance, the last segment of a plain network is much longer than the rest (see Appendix), which could bias the comparison with ResNet. We further define LSS by normalizing the length of each segment as follows
\begin{equation}
\tilde{x}^{(0)}=x^{(0)},\ 
\tilde{x}^{(l)}=\tilde{x}^{(l-1)}+\frac{x^{(l)}-x^{(l-1)}}{\|x^{(l)}-x^{(l-1)}\|_2},\ l=1,\cdots,n.\\
\end{equation}
\begin{equation}\label{score:lss}
\operatorname{LSS}=\frac{n}{\|\tilde{x}^{(n)}-\tilde{x}^{(0)}\|_2}.
\end{equation}
Obviously, $\operatorname{LSS}\geq1$, and when $\operatorname{LSS}=1$, the track is exactly located on a straight line. 

\par \textbf{Optimal transport score (OTS)}
We next compute the discrete optimal transport map from the input distribution to the output one, and test its consistency with the map learned by ResNets. Given a map $f$ of a ResNet, let $\tilde{y}_i=f(x_i)$ ($i=1,2,\cdots,m$). %, then $f$ is a map from data $\{x_i\}_{i=1}^{m}$ to $\{\tilde{y}_i\}_{i=1}^m$. 
Let $P_x$ and $P_y$ denote the empirical distributions of $\{x_i\}_{i=1}^{m}$ and $\{\tilde{y}_i\}_{i=1}^m$ respectively. The Wasserstein distance of $P_x$ and $P_y$ is
\begin{equation}
\begin{aligned}
W_2^2(P_x,P_y)&=\inf_{T_{\#}P_x=P_y}\int \|x-T(x)\|_2^2 dP_x(x)\\
&=\inf_{T_{\#}P_x=P_y}\frac{1}{m}\sum_{i=1}^m\|x_i-T(x_i)\|_2^2\\
&=\min_{\sigma}\frac{1}{m}\sum_{i=1}^{m}\|x_i-\tilde{y}_{\sigma(i)}\|_2^2,
\end{aligned}
\end{equation} 
where $\sigma$ is a permutation of an index set $\{1,2,\cdots,m\}$. It can be formulated as an assignment problem
\begin{equation}
\label{pb: assign problem}
\begin{aligned}
&\min_{c_{ij}}\ \sum_{i,j}c_{ij}\|x_i-\tilde{y}_j\|_2^2\\
&\mbox{s.t.}
\left\{\begin{aligned}
&\sum_{i=1}^{m}c_{ij}=1,\ j=1,2,\cdots,m,\\
&\sum_{j=1}^{m}c_{ij}=1,\ i=1,2,\cdots,m,\\
&c_{ij}\in\{0,1\},\\
\end{aligned}\right.
\end{aligned}
\end{equation}
where $c_{ij}=1$ means $x_i$ is transported to $\tilde{y}_j$, i.e., $\sigma(i)=j$. To measure the consistency of $f$ and $T$, we define the optimal transport score (OTS) as
\begin{equation}
\operatorname{OTS}=\frac{{\#}\{i\in\{1,2,\cdots,m\}|\sigma(i)=i\}}{m}.
\end{equation}
The range of OTS is $[0,1]$. A larger OTS indicates that the learned sample-wise pairing
$x_i \mapsto f(x_i)$ is more consistent with the discrete optimal assignment from $P_x$ to
the model-induced output distribution $P_y$. In particular, OTS = 1 means that the learned
pairing itself is an optimal assignment between these two empirical distributions. The assignment problem (\ref{pb: assign problem}) is solved using the Jonker-Volgenant algorithm \citep{jonker1987shortest}.

\par We compute the LSS during training and the OTS after training, both the plain network and ResNet, for classification on the MNIST, CIFAR-10, and CIFAR-100 datasets (Table 1). The plain networks in our experiments are set by removing the identity maps in the corresponding ResNets.

\begin{table*}[t!]
%	\tiny
%   \scriptsize
    \footnotesize
	\label{tab1}
	\renewcommand\arraystretch{1.6}
	\centering	
	\caption{Architectures of ResNet on MNIST, CIFAR-10 and CIFAR-100.}
	\begin{tabular}{c|c|c|c|c|c|c}
		\cline{1-7}
		Dataset&Data dim.& K & Intermediate dim. & No. of blocks & Layer type & Output dim. \\
		\cline{1-7}
		MNIST & 28$\times$28 & 1 & 1000 & 5 & Fully connected & 10 \\ 
		\cline{1-7}
		\multirow{5}{*}{CIFAR-10} & \multirow{5}{*}{3$\times$32$\times$32} & 1 & 32$\times$16$\times$16 & 10 & \multirow{5}{*}{Convolutional} & \multirow{5}{*}{10} \\   
		\cline{3-5}
		 &  & \multirow{4}{*}{4} & 64$\times$32$\times$32 & 5 &  &  \\   
		\cline{4-5}
		 &  &  & 128$\times$16$\times$16 & 5 &  &  \\ 
		\cline{4-5}
		&  &  & 256$\times$8$\times$8 & 5 &  &  \\   
		\cline{4-5}
		&  &  & 512$\times$4$\times$4 & 5 &  &  \\ 
		
		\cline{1-7}
		\multirow{5}{*}{CIFAR-100} & \multirow{5}{*}{3$\times$32$\times$32} & 1 & 32$\times$16$\times$16 & 10 & \multirow{5}{*}{Convolutional} & \multirow{5}{*}{100} \\   
		\cline{3-5}
		&  & \multirow{4}{*}{4} & 64$\times$32$\times$32 & 5 &  &  \\   
		\cline{4-5}
		&  &  & 128$\times$16$\times$16 & 5 &  &  \\ 
		\cline{4-5}
		&  &  & 256$\times$8$\times$8 & 5 &  &  \\   
		\cline{4-5}
		&  &  & 512$\times$4$\times$4 & 5 &  &  \\ 
		
		\cline{1-7}
		
	\end{tabular}
\end{table*}

\par \textbf{MNIST} The classification task on MNIST is easy to solve, such that both training and testing accuracy are almost $100\%$ under the following settings. We first increase the dimension of the data point from $28\times 28$ to 1000, and next use five ResNet blocks with fully connected layers to learn the transformation in $\mathbb{R}^{1000}$. Finally, we reduce the dimension from $1000$ to 10. 

\par \textbf{CIFAR-10 and CIFAR-100} For the classification tasks on CIFAR-10 and CIFAR-100, learning the transformation of distributions with the same input and output dimensions of $f$ is hard due to the redundant information in the data. As a consequence, the performance of both plain network and ResNet differs a lot with the same or varying dimensions, i.e., $K=1$ or $K>1$ in (\ref{eq:dimension_reduction}). Thus, we separate the numerical experiments for these two cases to test our principle. For $K=1$, we first increase the dimension from $3\times 32\times 32$ to $32\times 16\times 16$, and use 10 ResNet blocks with convolutional layers to learn the transformation of data representation on the $32\times16\times16$ tensor space. Finally, we reduce the dimensions from $32\times16\times16$ to 10 and 100, respectively. For $K=4$, we reduce the dimension of representation for better performance as follows:
\begin{equation}
64\times 32\times 32\to 128\times 16\times 16\to 256\times 8\times 8\to 512\times 4\times 4.
\end{equation}
For each dimension, we stack 5 ResNet blocks and compute the corresponding LSS and OTS, respectively.

\begin{figure*}[t!]	
	\centering
	\includegraphics[width=\columnwidth]{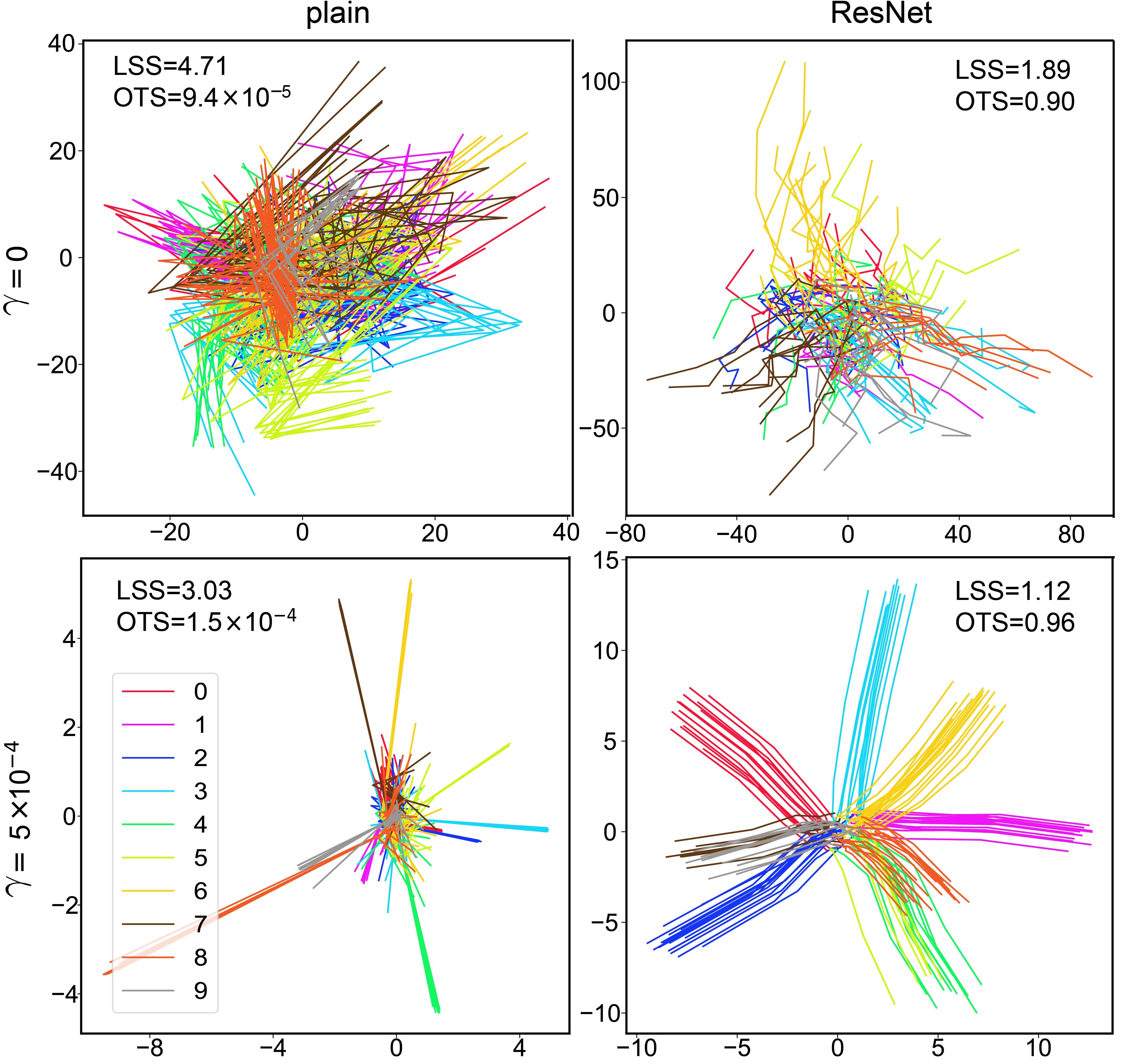}
	\caption{Two-dimensional visualization of data tracks with five segments for both plain networks and ResNets on MNIST. The original data representations of dimension 1000 are projected into two-dimensional space by multiplying them with a $2\times 1000$ random Gaussian matrix.}
	\label{fig3}
\end{figure*}
	
\begin{figure*}[t!]
	\centering
	\includegraphics[width=1.1\columnwidth]{figs/Fig18.jpg}
	\caption{Two-dimensional visualization of the trend of data tracks in the training progress. We present the data tracks of epochs 1, 40, 80, 120, and 160.}
	\label{fig4+}
\end{figure*}
\begin{figure*}[t!]
	\centering
	\includegraphics[width=\columnwidth]{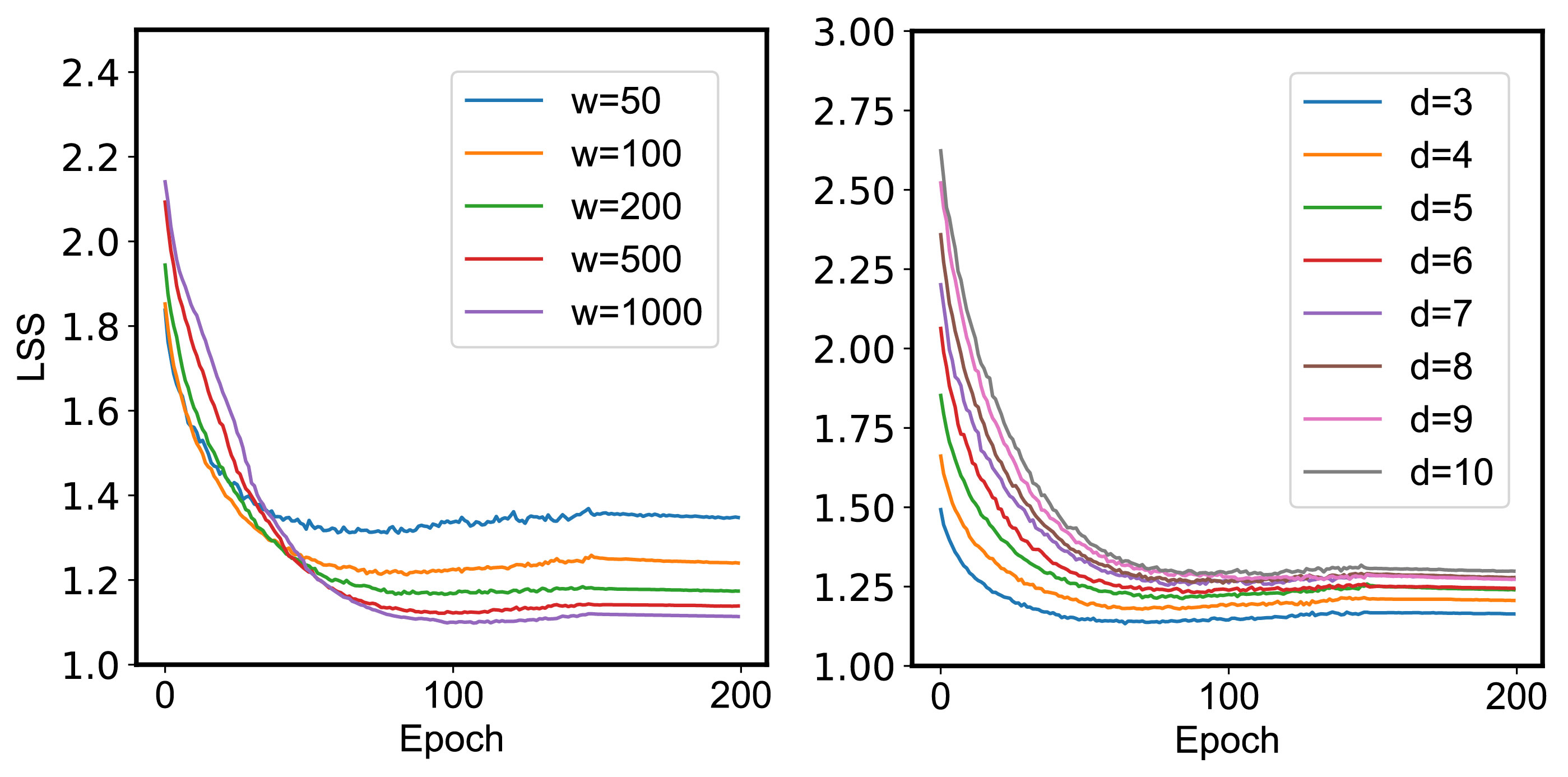}
	\caption{LSS scores of networks with different widths and depths. On the left, we fixed depth to 5 and tested networks with widths 50, 100, 200, 500, and 1000. On the right, we fixed the width to 100 and tested the depth from 3 to 10. }
	\label{fig5+}
\end{figure*}

\begin{figure*}[t!]
	\centering
	\includegraphics[width=\columnwidth]{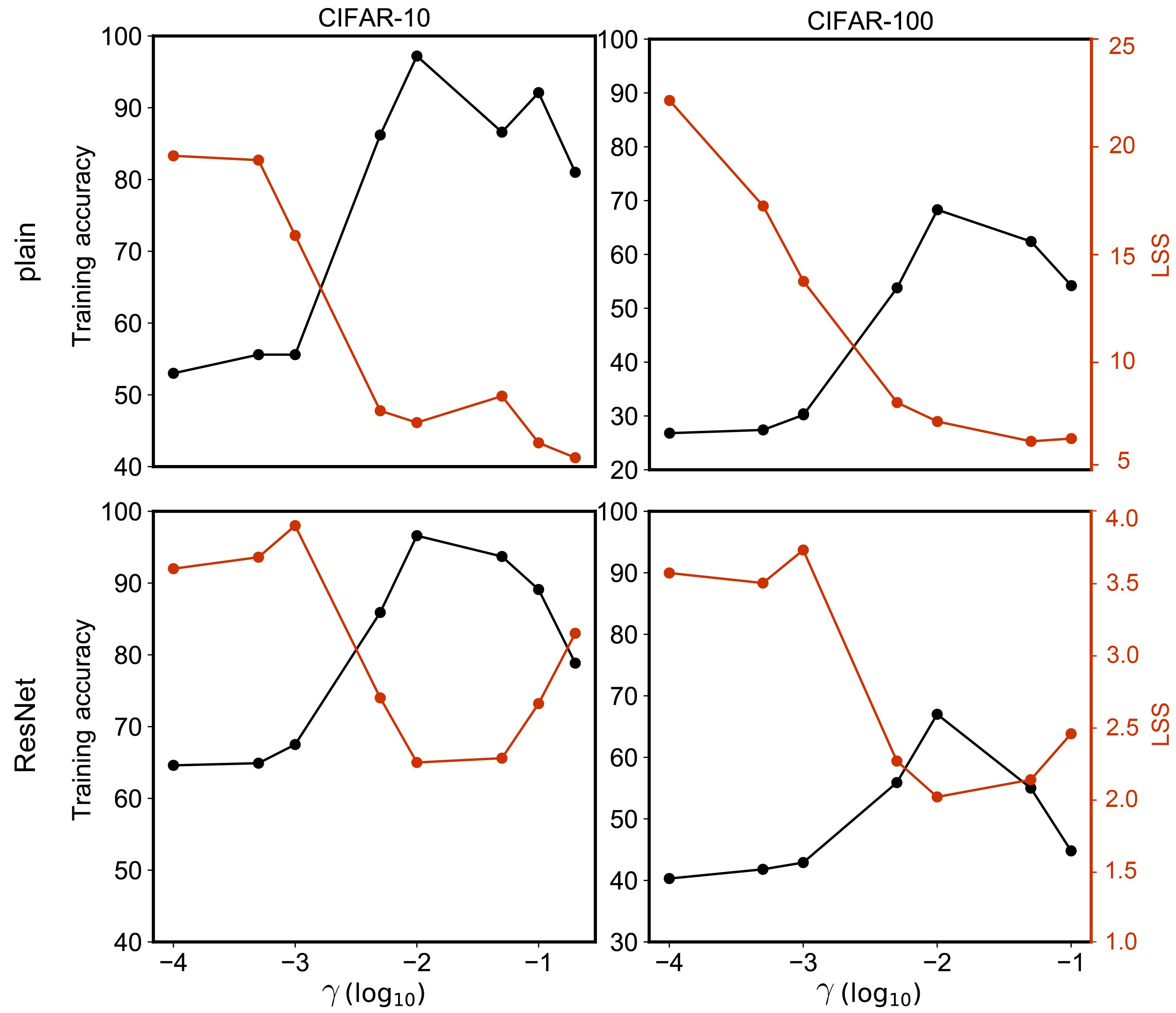}
	\caption{Comparison of change trend of training accuracy and LSS in terms of $\gamma$s for plain networks and ResNets with $K=1$ on CIFAR-10 and CIFAR-100, respectively.}
	\label{fig4}
\end{figure*}

\subsection{ResNets indeed tend to approximate the geodesic curve}
To verify that ResNets indeed tend to approximate the geodesic curve, we visualize the data tracks and compute the corresponding LSS and OTS. To visualize the data tracks, the original relatively high-dimensional data representations are projected into a two-dimensional space by multiplying them by a random Gaussian matrix. For MNIST, both plain network and ResNet could achieve very small LSS even with a relatively small $\gamma=5\times 10^{-4}$. 
While ResNet indeed results in line-shape data tracks ($\operatorname{LSS}=1.12$) and matches well with the optimal transport map ($\operatorname{OTS}=0.96$) with $\gamma=5\times 10^{-4}$, suggesting it could well approximate the geodesic curve in the Wasserstein space (Fig. \ref{fig3}).  We notice that the MNIST digital images of the same class are relatively similar, leading to many equations $y_i=f(x_i)$ to be solved being degenerate.

Thus, the tracks are more easily forced to straight lines by the weight decay. For CIFAR-10 and CIFAR-100, both plain network and ResNet require relatively large $\gamma$ to decrease LSS, and finally reached about 6.1 and 2.2, respectively (see Appendix). More figures for visualization and charts for LSS and OTS on CIFAR-10 and CIFAR-100 with $K=1$ or $K=4$ are presented in the Appendix, demonstrating that ResNets indeed tend to approximate the geodesic curve under various settings. We show the visualization trend of data tracks of different epochs (Fig. \ref{fig4+}), which indicates that ResNet gradually approximates the geodesic curve during training. We also test LSS of ResNets with different widths and depths (Fig. \ref{fig5+}). We found that, under different settings, the tracks uniformly approached the straight lines during training, and the LSS decreased as the width increased.

\subsection{Better line shape of data tracks leads to higher accuracy}
In Section 3, we derive heuristically that a larger $\gamma$ can enhance the optimization performance of ResNets by enforcing it to approximate the geodesic curve. To verify this, we test both plain network and ResNet with $K=1$ and compare the variation trend of training accuracy and LSS with respect to multiple $\gamma$s (Fig. \ref{fig4}). According to our experiments, LSS and the training accuracy are highly correlated. We observe that the training accuracy increases when the LSS decreases dramatically with $\gamma\in [10^{-3},10^{-2}]$ on both CIFAR-10 and CIFAR-100, respectively. It demonstrated that, when the data tracks are closer to straight lines, both networks can better fit the training data, which strongly supports our theoretical derivation about optimization. When $\gamma$ is greater than 0.1, the weight decay tends to hinder the fitting process, and both training and test accuracy will decline.

\begin{figure}[h!]	
	\centering
	\includegraphics[width=0.95\columnwidth]{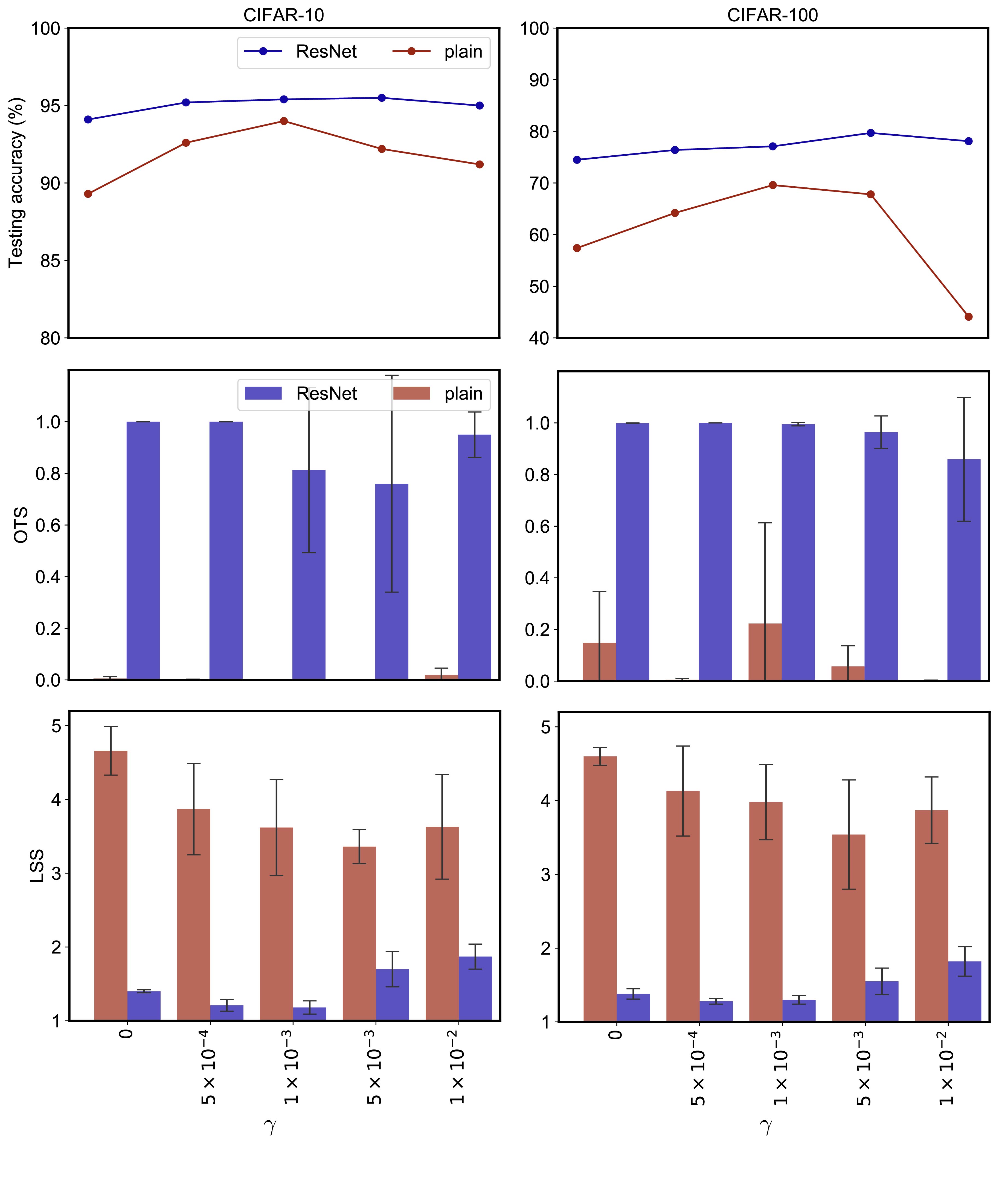}
	\caption{Performance comparison of plain networks and ResNets in terms of testing accuracy, OTS, and LSS for four diverse dimensions (i.e., $K=4$) on CIFAR-10 and CIFAR-100, respectively. Four hyperparameters $\gamma$s are used for evaluation.  }
	\label{fig6}
\end{figure}

\subsection{Approximating the geodesic curve leads to better generalization}
In Section 4, we conclude that approximating the geodesic curve leads to better generalization for ResNets by lower-bounding the distance of data tracks. To verify it, we test both plain network and ResNet with $K=4$ on CIFAR-10 and CIFAR-100 since the training accuracy under such cases is close to 100$\%$. When reducing the dimensions to $K=4$, the information irrelevant to the label can be dropped. Thus, the classification problem becomes relatively easy to solve, and the training accuracy of ResNets and plain networks is close to $100\%$ on both CIFAR-10 and CIFAR-100, respectively. 
Both plain network and ResNet with $K=4$ can obtain smaller LSSs than those with $K=1$, i.e., about 4 and 1.2 for each dimension, respectively (see Appendix).  
Even though the performance of ResNets and plain networks is similar on training data, ResNets demonstrate superior test accuracy to plain networks. We can clearly see that the OTS and LSS of ResNets are closer to 1 (Fig. \ref{fig6}), which suggests that ResNets approximate the geodesic curve better than plain networks. 
On the contrary, the tracks of plain networks zigzag with high LSS and are arranged with OTS closing to 0. 
It illustrates why ResNets generalize better in practice. We also confirm the derivations that ResNets are not dominated by individual units and are more robust to both random and adversarial noise (see Appendix). The variance of scores on different dimensions in both ResNets and plain networks is large. This is due to an imbalance in the number of parameters in different dimensions. For instance, the $64\times 32 \times 32$ tensor is 8 times the dimension of the $512\times 4\times 4$ tensor; however, while utilizing $3\times 3$ convolution kernels, the number of parameters ($512\times 512 \times 3\times3$) in each layer on the latter is 64 times that of the former ($64\times 64\times 3\times 3$). Therefore, the tracks on the $512\times 4\times 4$ dimension are more flexible to be linearized than those on the $64\times 32 \times 32$ dimension. We can conjecture that the performance of DNN can be influenced by this kind of imbalance, which may increase with $\gamma$. As a consequence of that, on the dimension-varying situation $K=4$, the best testing accuracy of ResNet and plain network is achieved when $\gamma=5\times10^{-3}$ and $\gamma=1\times10^{-3}$ respectively, while it is achieved with $\gamma=1\times10^{-2}$ for $K=1$.

\section{Conclusion and discussion}
\par Deep learning has achieved remarkable success across vision, language, scientific computing, and many other domains. However, its internal mechanism is still far from fully understood. In this paper, we interpret the forward propagation of DNNs as a measure-preserving evolution governed by continuity equations, and connect the learned layerwise transformation to geodesic curves in the Wasserstein space. From this viewpoint, a network is not merely a composition of nonlinear maps, but a discretized transport process that moves the input distribution toward a task-dependent output distribution.

\par Our results also connect with recent studies on representation geometry. In particular, the neural collapse phenomenon shows that, during the terminal phase of training, within-class features collapse to their class means, and the class means converge to a simplex equiangular tight frame \citep{papyan2020prevalence}. Combining this phenomenon with the observations in this paper suggests an idealized model of deep residual networks for classification: the network transports the data distribution, approximately along a Wasserstein geodesic, toward a class-structured simplex ETF. This perspective is consistent with recent studies on progressive feedforward collapse, which show that intermediate representations in ResNets can progressively collapse toward class-structured geometry during training \citep{wang2024progressive}. Related geometric phenomena, including residual alignment and layer-wise representation similarity, have also been explored in recent work \citep{li2023residual}\citep{jiang2025tracing}. Beyond ResNets, trajectory-based geometry remains relevant for Transformer-based models: Transformer block coupling studies token trajectories in open-source LLMs and shows that coupling and trajectory linearity correlate with model performance \citep{aubry2025transformer}. How to characterize the contribution of each layer to the final prediction in large language models and specialized deep models, from both statistical and semantic perspectives, remains an important question for future research.

\par Another promising direction is to use the intrinsic connection between residual networks and optimal transport maps to improve network design. Under suitable assumptions, optimal transport maps enjoy favorable regularity properties, such as local Lipschitz continuity. It is especially relevant because neural networks are known to be sensitive to adversarial perturbations. \citet{gai2024otad} explored this direction by deriving an optimal-transport-induced regularizer and solving a convex integration problem so that the learned model preserves local Lipschitz properties and becomes more robust to agnostic adversarial attacks. How to further exploit the connection between optimal transport and residual architectures to design more stable, robust, and interpretable deep networks is also an important direction for future work.

\acks{This work has been supported by the CAS Project for Young Scientists in Basic Research [No. YSBR-034 to S.Z.], the Strategic Priority Research Program of the Chinese Academy of Sciences [No. XDB0680101 to S.Z.], the National Natural Science Foundation of China [Nos. 32341013, 12326614], and the Robotic AI-Scientist Platform of the Chinese Academy of Sciences.}
	
\vskip 0.2in
\bibliography{reference}

\clearpage
\section{Appendix}

\subsection{ResNet is not dominated by individual units}
Some studies \citep{zhou2018revisiting}\citep{bau2020understanding}\citep{bau2018identifying} revealed that each individual unit of DNN has different semantics and matches a diverse set of object concepts. As we discussed, the data tracks of plain network zigzag to change their directions frequently, and data representations of ResNet are approximately forward-propagated through line-shaped tracks. Thus, for ResNet, each layer contributes a part to the final transportation without significantly changing the direction of tracks. 

\par
\begin{figure}[h!]
	\centering
	\includegraphics[width=0.55\columnwidth]{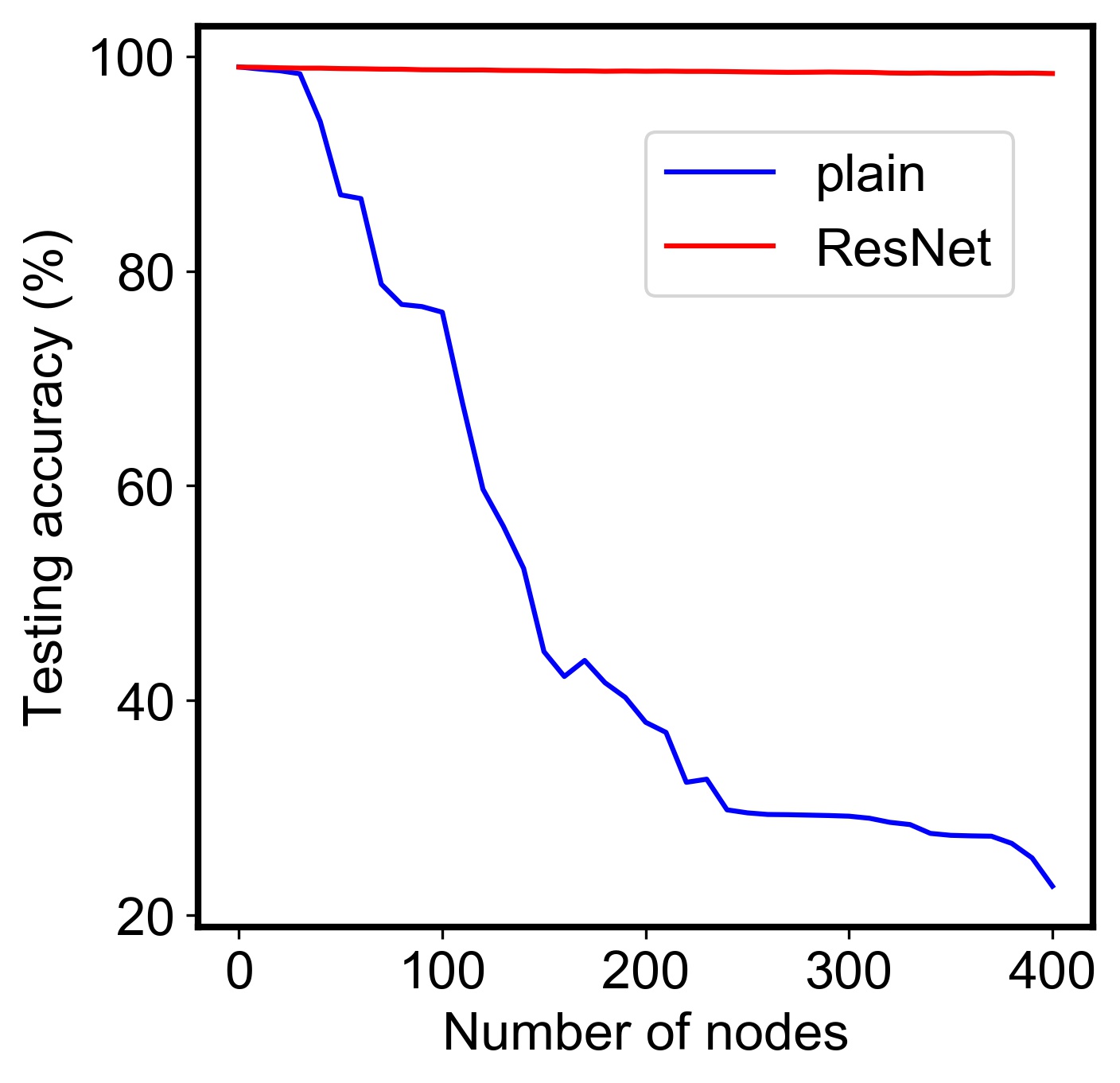}
	\caption{Testing accuracy of the five-block ResNet and corresponding plain network with the eliminated units in the 3rd layer on MNIST.}
	\label{fig7}
\end{figure}

 To test the contribution of individual units in the $l$-th layer, we first obtain the most important units corresponding to each class by the average of all the data points in the class. Then for each $x$, we set the important units of $x^{(l)}$ to the same as that of $x^{(l-1)}$ to eliminate the effect of the $l$-th layer. Experiments show that the testing accuracy of ResNet doesn't change significantly. But that of the plain network decreases rapidly with the increase in the number of eliminated units (Fig. \ref{fig7}). This observation confirms our theoretical derivation that ResNet is not dominated by individual units, since it approximates the geodesic curve in the Wasserstein space. 

\subsection{ResNet exhibits improved robustness to random and adversarial perturbations}
\begin{figure}[b!]
	\centering
	\includegraphics[width=0.90\columnwidth]{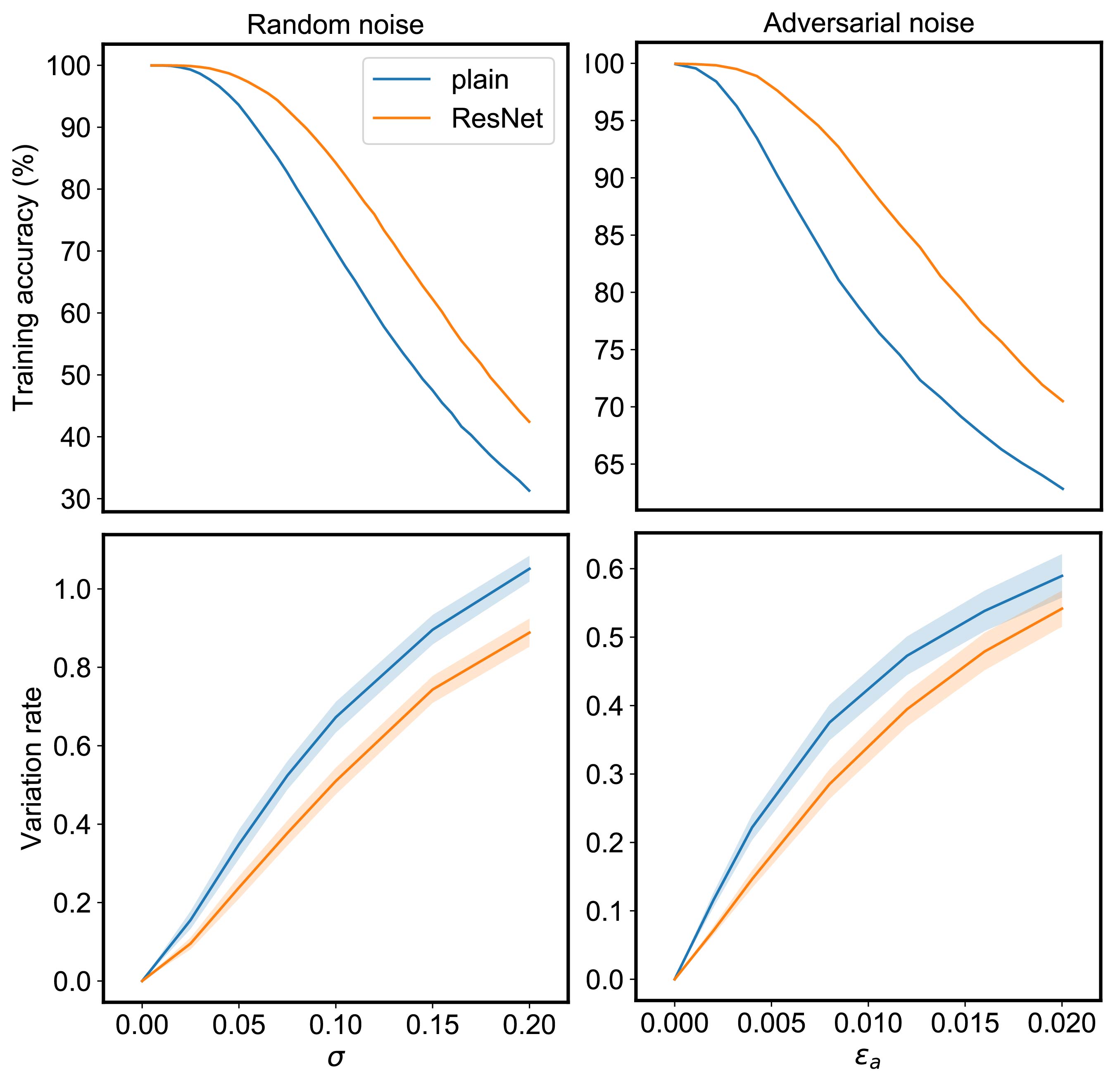}
	\caption{Comparison of plain network and ResNet on the data with random noise (left) and adversarial noise (right).}
	\label{fig8}
\end{figure}

According to the theoretical analysis in Section 4, ResNet is more robust to noise, since it better approximates the geodesic curve in the Wasserstein space. To test this, we train a ResNet and a plain network with $K=4$ on CIFAR-100. After training, the loss over all the training data for the ResNet and plain network is 2.075 and 1.773, respectively. However, numerical experiments show that ResNet is much more robust to both random and adversarial noise, and the classification accuracy falls more slowly than a plain network with an increase in noise (Fig. \ref{fig8}). 

\par For each data point $x$ in the dataset $\mathcal{D}$, the data with random noise can be denoted by $x_r=x+e_r$, where $e_r\sim \mathcal{N}(0_{d\times 1},\sigma)$. The data with adversarial noise is constructed by the fast gradient sign method (FGSM) \citep{goodfellow2014explaining}, i.e.,
$x_a=x+\epsilon_a\operatorname{sign}\left(\frac{\partial\mathcal{L}}{\partial x}\right)$, where $\epsilon_a$ is the coefficient of the noise and sign indicates the positive or negative symbol. Then the variation rate in the $l$-th layer by adding random or adversarial noise to the original data $\operatorname{Vr}_r(l,\sigma)$ and $\operatorname{Vr}_a(l,\sigma)$ can be defined by 
\begin{equation}
\operatorname{Vr}_r(l,\sigma)=\sum_{x\in \mathcal{D}}\frac{\|x^{(l)}-x^{(l)}_r\|_2}{\|x^{(l)}\|_2|\mathcal{D}|},\quad\operatorname{Vr}_a(l,\sigma)=\sum_{x\in \mathcal{D}}\frac{\|x^{(l)}-x^{(l)}_a\|_2}{\|x^{(l)}\|_2|\mathcal{D}|}.
\end{equation} 
As we can see, the variation rate curve of the ResNet when $l=n$ is lower than that of the plain network (Fig. \ref{fig8}), indicating that ResNet indeed decreases the probability for the noisy data to be misclassified.

\subsection{Proportion of segment lengths of data tracks of plain networks and ResNets on CIFAR-100}
We show the proportion of segment lengths of data tracks of plain networks and ResNets on CIFAR-100 with $K=4$, respectively (Tables 2 and 3).

\begin{table*}[h!]
	\tiny
	\label{tab6}
	\renewcommand\arraystretch{1.8}
	\centering	
	\caption{Proportion of segment lengths of data tracks in plain networks on CIFAR-100.}
	\begin{tabular}{c|c|c|c|c|c|c|c|c|c|c}
		\cline{1-11}
		$\gamma$ & $S_1$ & $S_2$ & $S_3$ & $S_4$ & $S_5$ & $S_6$ & $S_7$ & $S_8$ & $S_9$ & $S_{10}$ \\
		
		\cline{1-11}
		0 & 0.076 & 0.103 & 0.093 & 0.104 & 0.106 & 0.102 & 0.105 & 0.107 & 0.117 & 0.087 \\ 
		\cline{1-11}
		$1\times10^{-4}$ & 0.079 & 0.097 & 0.103 & 0.097 & 0.103 & 0.103 & 0.107 & 0.112 & 0.115 & 0.083 \\   
		\cline{1-11}
		$5\times10^{-4}$ & 0.072 & 0.102 & 0.103 & 0.106 & 0.101 & 0.114 & 0.099 & 0.119 & 0.099 & 0.086 \\   
		\cline{1-11}
		$1\times10^{-3}$ & 0.076 & 0.105 & 0.099 & 0.109 & 0.098 & 0.092 & 0.094 & 0.102 & 0.112 & 0.107\\   
		\cline{1-11}
		$5\times10^{-3}$ & 0.035 & 0.039 & 0.042 & 0.041 & 0.039 & 0.042 & 0.039 &  0.038 & 0.036 & 0.643\\   
		\cline{1-11}
		$1\times10^{-2}$ & 0.019 & 0.012 & 0.013 & 0.013 & 0.013 & 0.012 & 0.012 &  0.011 & 0.010 & 0.880\\   
		\cline{1-11}
		$5\times10^{-2}$ & 0.010 & 0.003 & 0.003 & 0.003 & 0.003 & 0.002 &
		0.002 & 0.001 & 0.001 &  0.969  \\   
		\cline{1-11}
		$1\times10^{-1}$ & 8.97$\times 10^{-3}$ & 2.95$\times 10^{-3}$ & 2.85$\times 10^{-3}$ & 2.55$\times 10^{-3}$ & 1.78$\times 10^{-3}$ & 1.40$\times 10^{-3}$ & 1.05$\times 10^{-3}$ & 6.92$\times 10^{-4}$ & 8.77$\times 10^{-4}$ & 0.977 \\   
		\cline{1-11}
		$5\times10^{-1}$ & 0.118 & 0.130 & 0.109 & 0.093 & 0.088 & 0.103 & 0.091 & 0.087 & 0.080 &  0.095   \\   
		\cline{1-11}
		
	\end{tabular}
\end{table*}

\begin{table*}[h!]
	\scriptsize
	\label{tab7}
	\renewcommand\arraystretch{1.8}
	\centering	
	\caption{Proportion of segment lengths of data tracks in ResNets on CIFAR-100.}
	\begin{tabular}{c|c|c|c|c|c|c|c|c|c|c}
		\cline{1-11}
		$\gamma$ & $S_1$ & $S_2$ & $S_3$ & $S_4$ & $S_5$ & $S_6$ & $S_7$ & $S_8$ & $S_9$ & $S_{10}$ \\
		\cline{1-11}
		0 &0.264 & 0.121 & 0.088 & 0.087 & 0.085 & 0.074 & 0.067 & 0.059 & 0.082 & 0.068 \\ 
		\cline{1-11}
		$1\times10^{-4}$ &0.218 & 0.126 & 0.091 & 0.083 & 0.076 & 0.075 & 0.082 & 0.071 & 0.091 & 0.084 \\   
		\cline{1-11}
		$5\times10^{-4}$ & 0.238 & 0.111 & 0.109 & 0.074 & 0.066 & 0.091 & 0.075 & 0.088 & 0.068 & 0.075 \\   
		\cline{1-11}
		$1\times10^{-3}$ & 0.247 & 0.111 & 0.092 & 0.090 & 0.069 & 0.084 & 0.060 & 0.070 & 0.091 & 0.082\\   
		\cline{1-11}
		$5\times10^{-3}$ & 0.067 & 0.059 & 0.074 & 0.068 & 0.087 & 0.092 & 0.084 & 0.135 & 0.147 & 0.183\\   
		\cline{1-11}
		$1\times10^{-2}$ & 0.009 & 0.007 & 0.009 & 0.010 & 0.029 & 0.079 & 0.204 & 0.184 & 0.242 & 0.223 \\   
		\cline{1-11}
		$5\times10^{-2}$ & 0.002 & 0.002 & 0.002 & 0.003 & 0.009 & 0.171 & 0.219 & 0.182 & 0.239 &  0.168  \\   
		\cline{1-11}
		$1\times10^{-1}$ & 0.001 & 0.001 & 0.001 & 0.002 & 0.002 & 0.003 & 0.012 & 0.203 & 0.054 & 0.721 \\   
		\cline{1-11}
		$5\times10^{-1}$ &0.145 & 0.097 & 0.090 & 0.102 & 0.097 & 0.085 & 0.099 & 0.093 & 0.111 & 0.078  \\   
		\cline{1-11}
		
	\end{tabular}
\end{table*}

\clearpage
\subsection{Visualization of data tracks of plain networks and ResNets}
We show data tracks of plain networks and ResNets with $K=1$ and $K=4$ on both CIFAR-10 and CIFAR-100, respectively (Figures 9-16).
\begin{figure}[h!]
	\centering
	\includegraphics[width=\columnwidth]{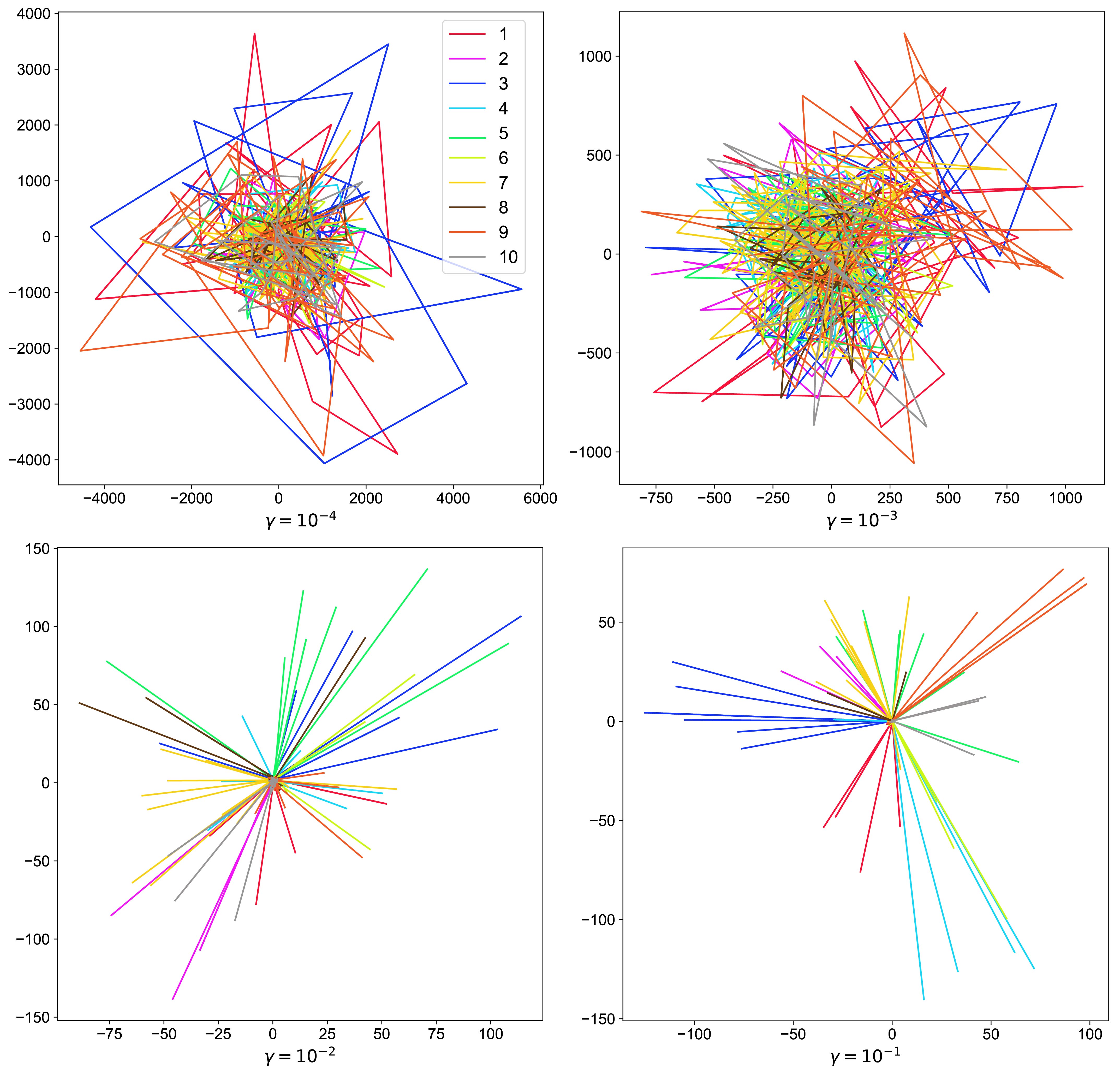}
	\caption{Visualization of data tracks of plain networks when $K=1$ with $\gamma=10^{-4}$, $\gamma=10^{-3}$, $\gamma=10^{-2}$, $\gamma=10^{-1}$ on CIFAR-10.}
	%\label{fig: noisy data}
\end{figure}

\begin{figure}[h!]
	\centering
	\includegraphics[width=0.95\columnwidth]{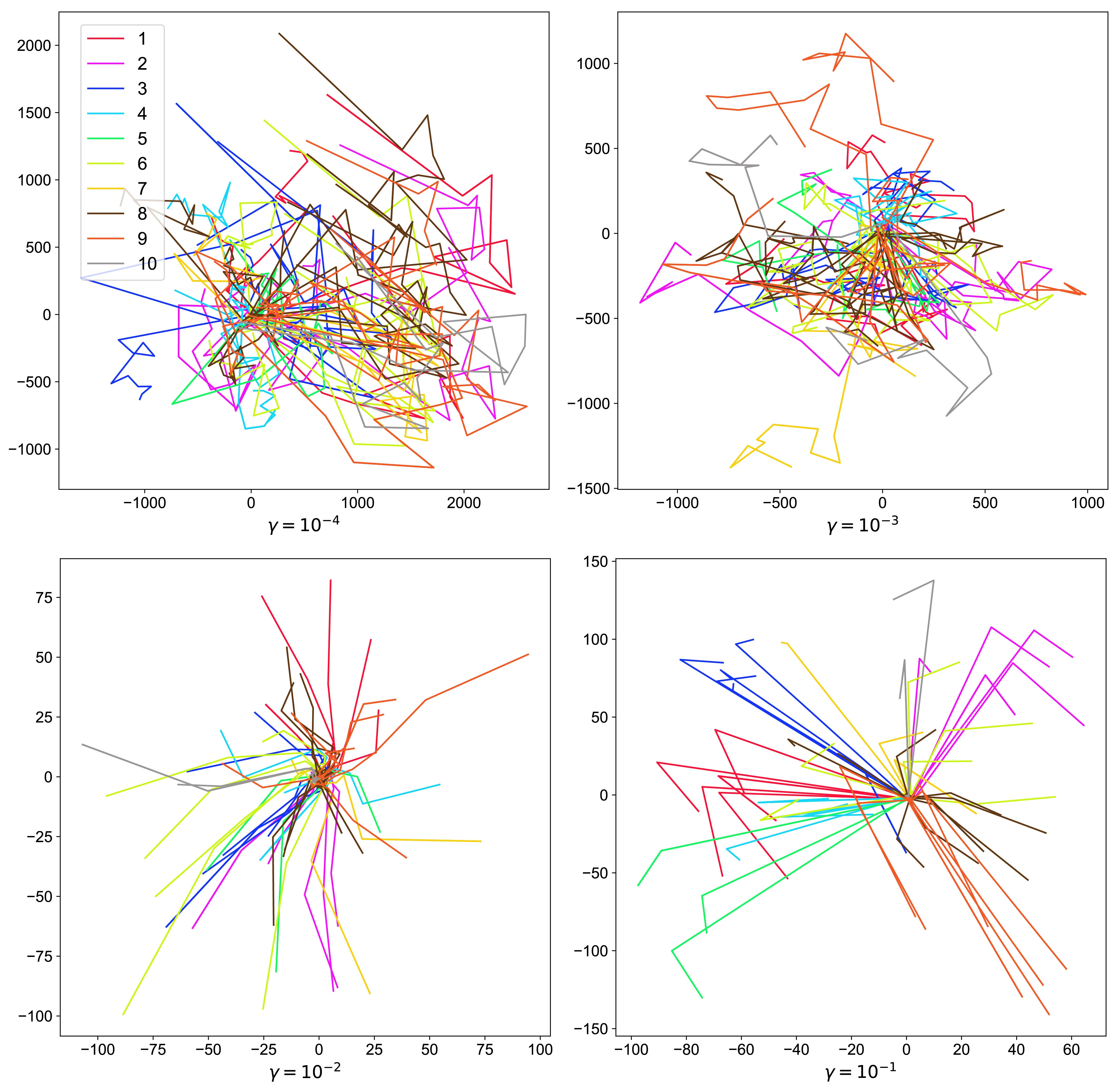}
	\caption{Visualization of data tracks of ResNets when $K=1$ with $\gamma=10^{-4}$, $\gamma=10^{-3}$, $\gamma=10^{-2}$, $\gamma=10^{-1}$ on CIFAR-10.}
	%\label{fig: noisy data}
\end{figure}
%%%%%%%%%%%%%%%%%%%%%%%%%%%%%%%%%%%%%%%%%

\begin{figure}[h!]	
	\centering
	\includegraphics[width=\columnwidth]{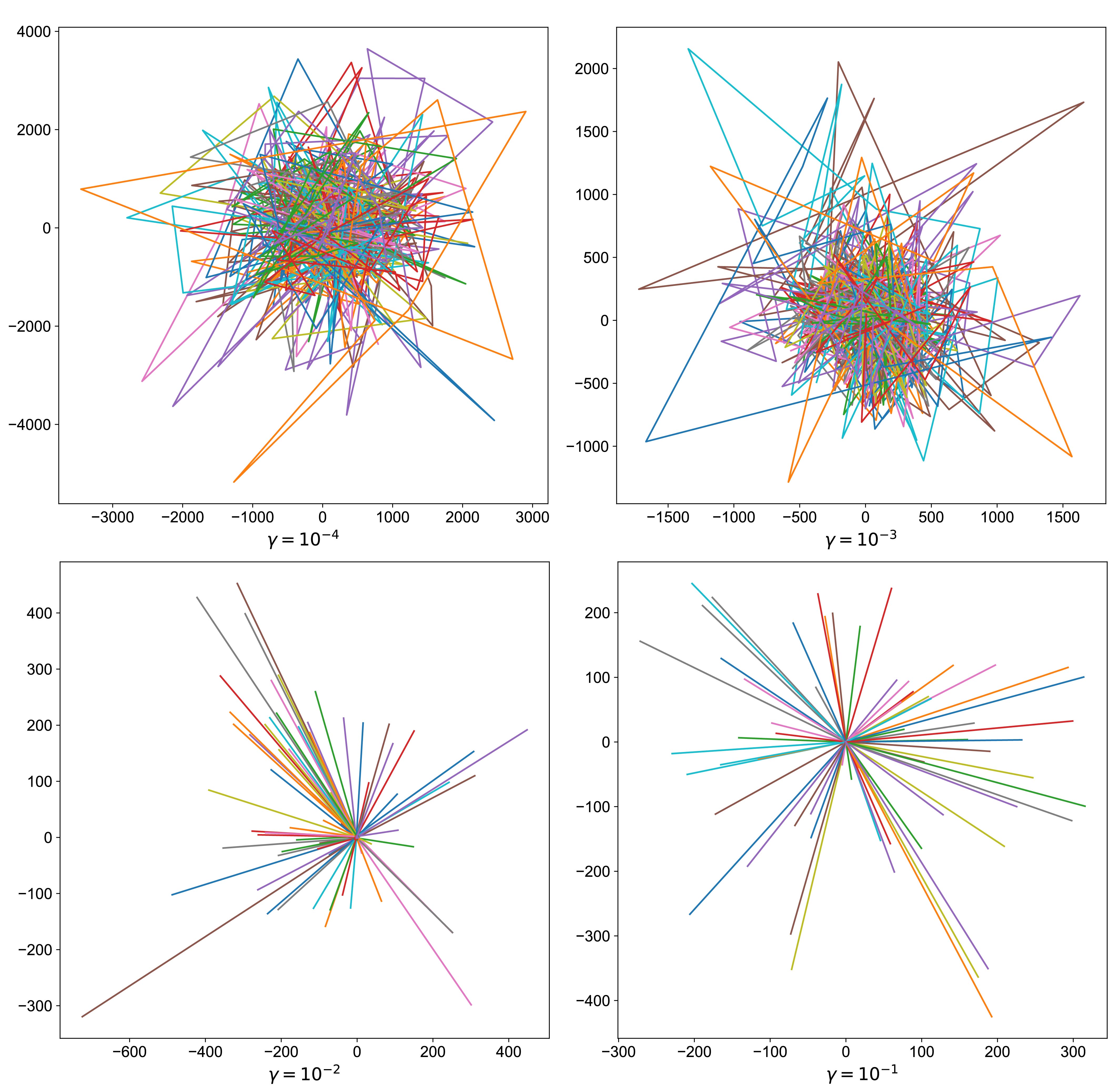}
	\caption{Visualization of data tracks of plain networks when $K=1$ with $\gamma=10^{-4}$, $\gamma=10^{-3}$, $\gamma=10^{-2}$, $\gamma=10^{-1}$ on CIFAR-100.}
	%\label{fig: noisy data}
\end{figure}	

\begin{figure}[h!]
	\centering
	\includegraphics[width=\columnwidth]{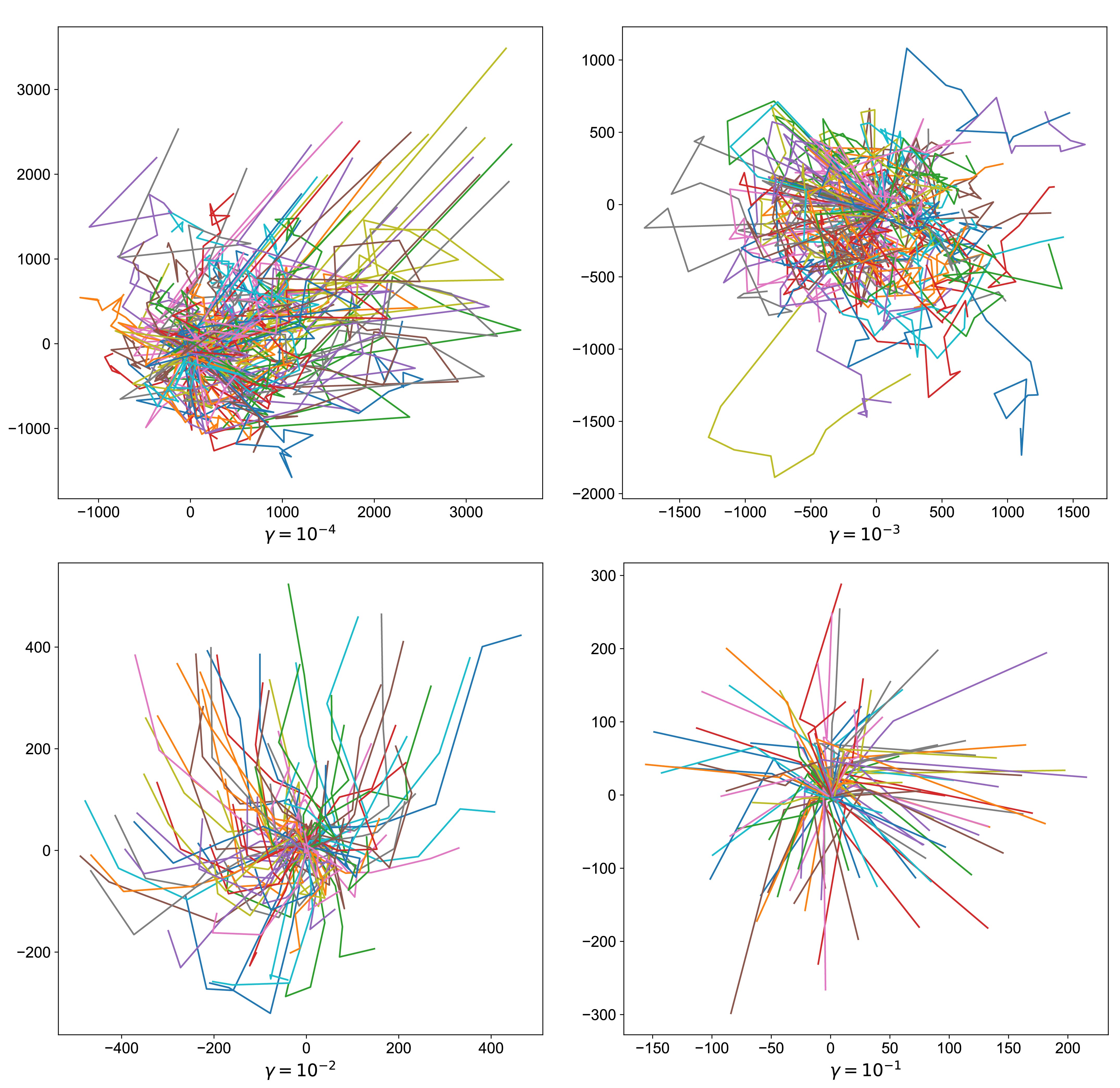}
	\caption{Visualization of data tracks of ResNets when $K=1$ with $\gamma=10^{-4}$, $\gamma=10^{-3}$, $\gamma=10^{-2}$, $\gamma=10^{-1}$ on CIFAR-100.}
	%\label{fig: noisy data}
\end{figure}
%%%%%%%%%%%%%%%%%%%%%%%%%%%%%%%%%%%%%%%%%

\begin{figure}[h!]
	\centering
	\includegraphics[width=\columnwidth]{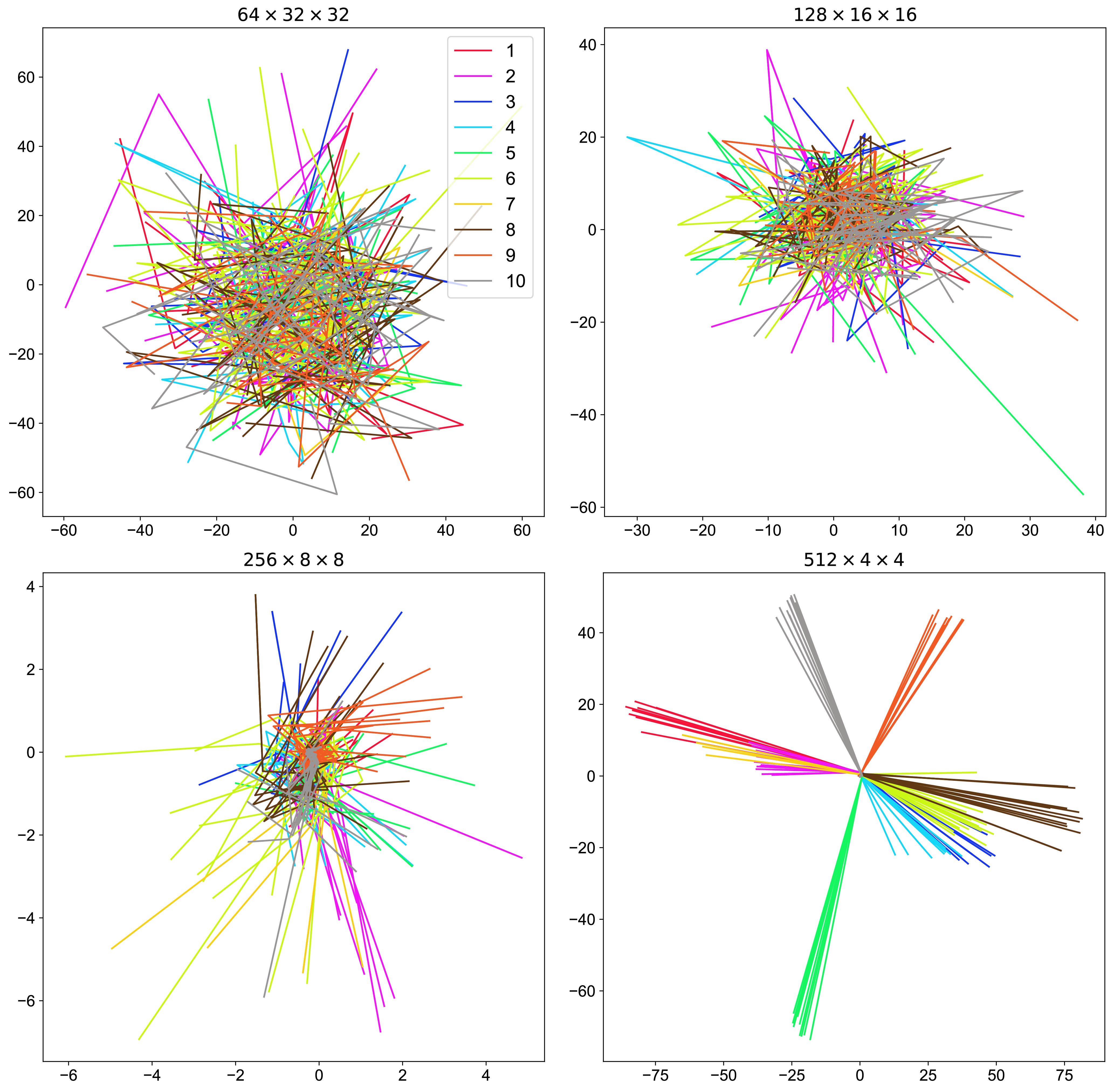}
	\caption{Visualization of data tracks of plain networks when $K=4$ with $\gamma=10^{-3}$ on CIFAR-10.}
	%\label{fig: noisy data}
\end{figure}

\begin{figure}[h!]
	\centering
	\includegraphics[width=\columnwidth]{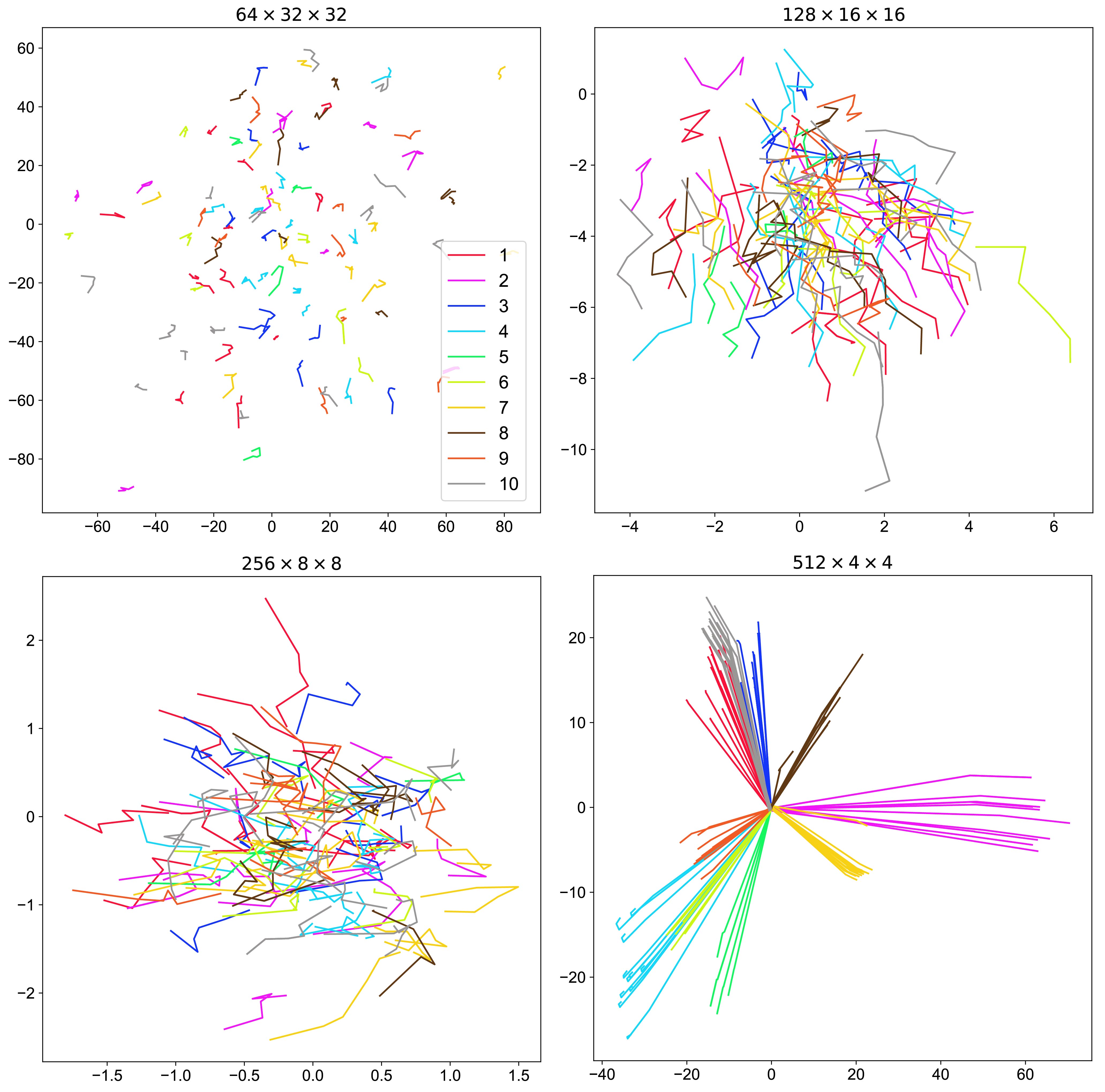}
	\caption{Visualization of data tracks of ResNets when $K=4$ with $\gamma=5\times 10^{-3}$ on CIFAR-10.}
	%\label{fig: noisy data}
\end{figure}
%%%%%%%%%%%%%%%%%%%%%%%%%%%%%%%%%%%%%%%%%

\begin{figure}[h!]	
	\centering
	\includegraphics[width=\columnwidth]{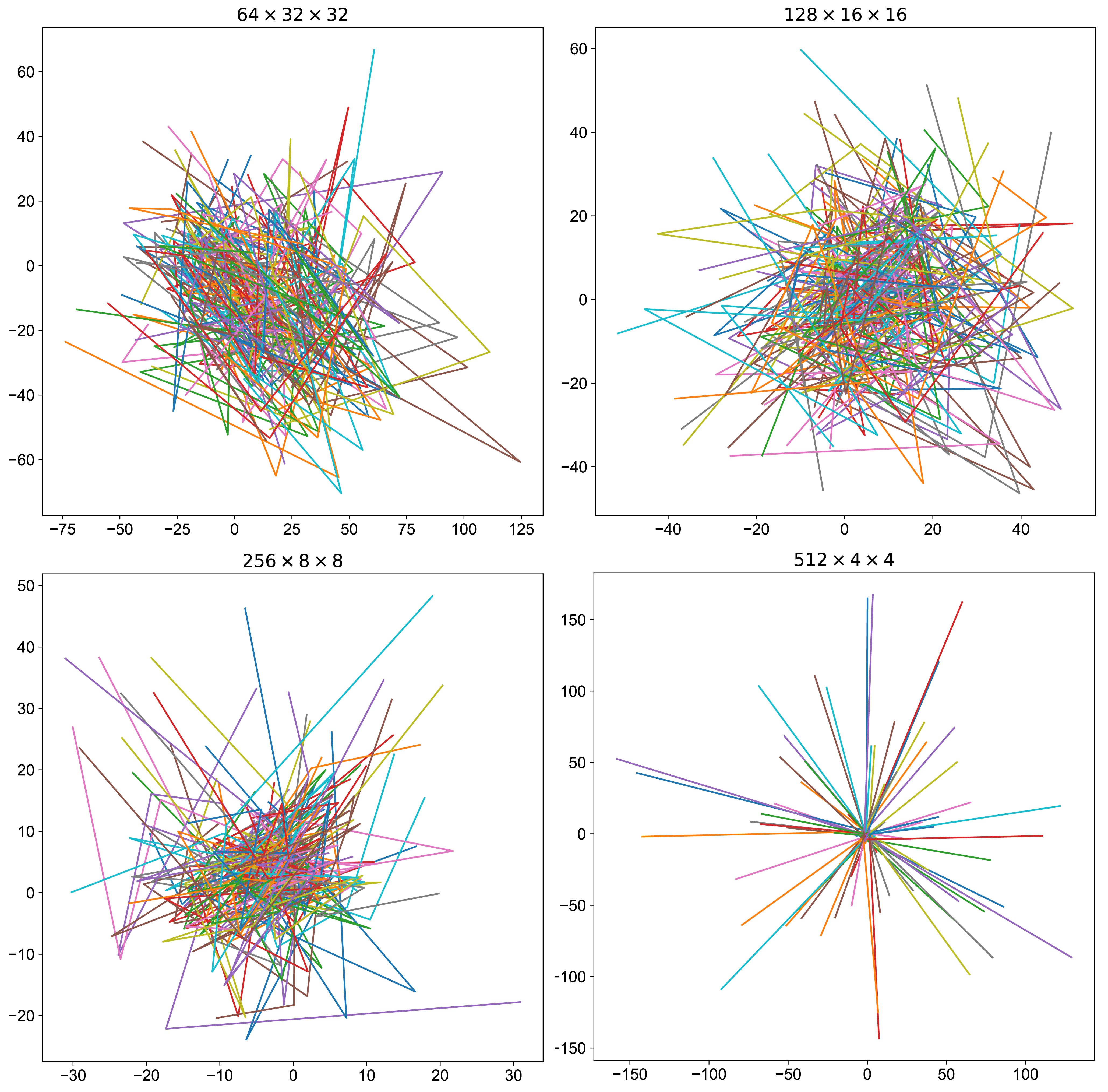}
	\caption{Visualization of data tracks of plain networks when $K=4$ with $\gamma=10^{-3}$ on CIFAR-100.}
	%\label{fig: noisy data}
\end{figure}

\begin{figure}[h!]	
	\centering
	\includegraphics[width=\columnwidth]{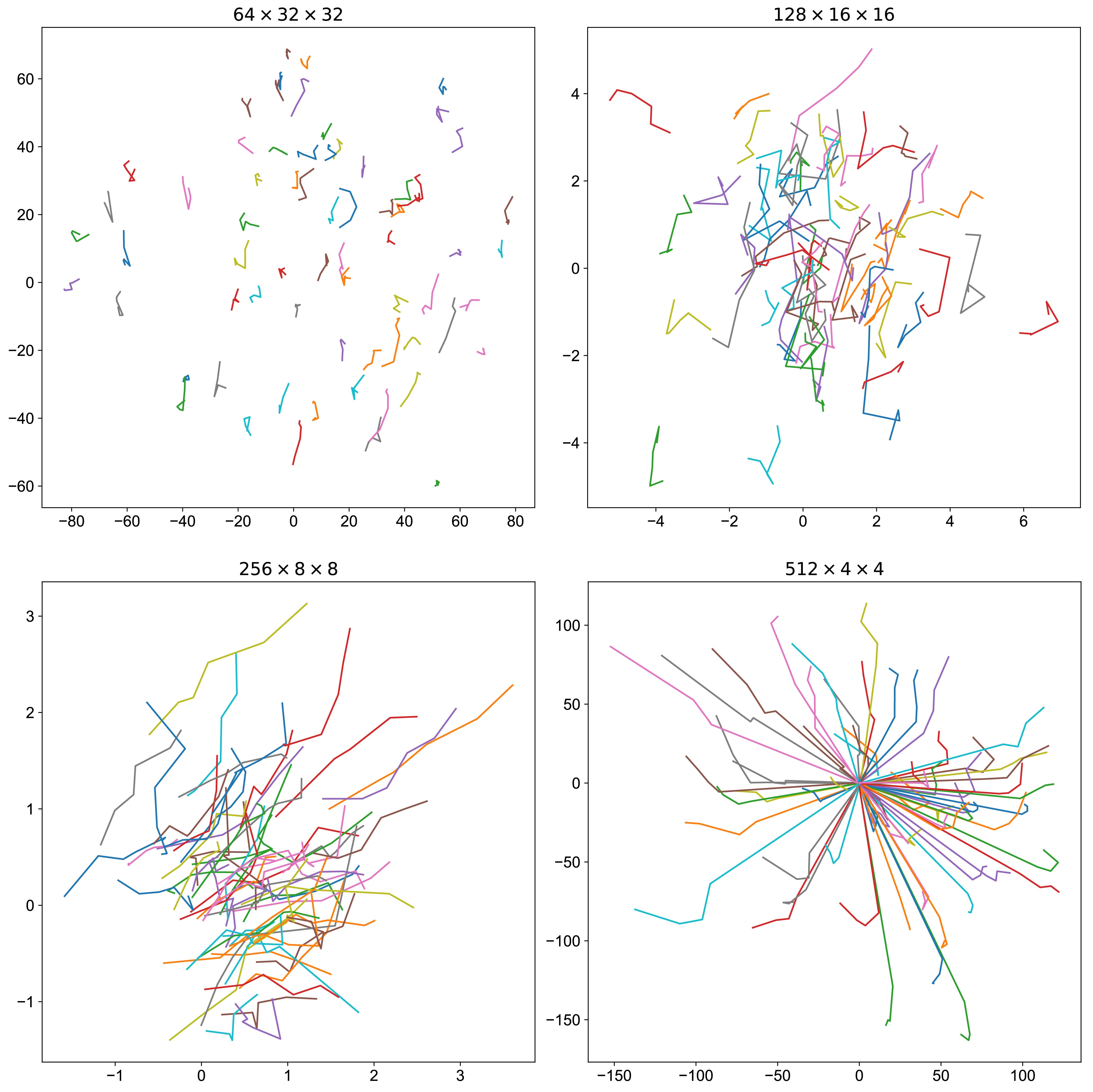}
	\caption{Visualization of data tracks of ResNets when $K=4$ with $\gamma=5\times10^{-3}$ on CIFAR-100.}
	%\label{fig: noisy data}
\end{figure}
	
%%%%%%%%%%%%%%%%%%%%%%%%%%%%%%%%%%%%%%%%%%%%%

\clearpage
\subsection{Performance comparison between plain networks and ResNets}
We demonstrate the performance comparison between plain networks and ResNets with $K=1$ and $K=4$ in terms of testing accuracy, OTS, LSR, and LSS on CIFAR-10 and CIFAR-100, respectively (Tables 4-7).
\begin{table*}[h!]
	\tiny %\scriptsize
	\renewcommand\arraystretch{1.8}
	\centering	
	\caption{Performance comparison of plain networks and ResNets with $K=1$ in terms of testing accuracy, OTS, LSR, and LSS on the CIFAR-10.}
	\begin{tabular}{c|c|c|c|c|c|c|c|c|c|c}
		\cline{1-11}
		\multirow{2}*{$\gamma$}&\multicolumn{2}{c|}{Training Accuracy ($\%$)}&\multicolumn{2}{c|}{Testing Accuracy ($\%$)}&\multicolumn{2}{c|}{OTS} &\multicolumn{2}{c|}{LSR}&\multicolumn{2}{c}{LSS}\\
		\cline{2-11}
		~& plain & ResNet & plain & ResNet & plain & ResNet & plain & ResNet& plain & ResNet \\
		\cline{1-11}
		0 & 52.1 & 63.0 & 53.5 & 66.3 & $9.0\times 10^{-4}$ & 0.681 & 78.5 & 3.71 & 20.2 & 3.79 \\
		\cline{1-11}
		$1\times 10^{-4}$ & 53.0 & 64.6 & 54.9 & 67.5 & $2.3\times 10^{-3}$ & 0.329 & 67.9 & 3.30 & 19.1 & 3.60 \\
		\cline{1-11}
		$5\times 10^{-4}$ & 55.6 & 64.9 & 57.0 & 67.7 & $2.1\times 10^{-3}$ & 0.753 & 51.2 & 3.57 & 18.9 & 3.68 \\
		\cline{1-11}
		$1\times 10^{-3}$ & 55.6 & 67.5 & 57.0 & 70.3 & $1.0\times 10^{-3}$ & 0.826 & 28.8 & 3.81 & 15.4 & 3.90 \\  
		\cline{1-11}
		$5\times 10^{-3}$ & 86.2 & 85.9 & 82.4 & 82.9 & $8.0\times 10^{-4}$ & 0.0242 & 2.26 & 2.07 & 7.26 & 2.70 \\
		\cline{1-11}
		$1\times 10^{-2}$ & 97.2 & 96.6 & 88.9 & 88.5 & $9.0\times 10^{-4}$ & 0.0119 & 1.27 & 1.24 & 6.71 & 2.25 \\
		\cline{1-11}
		$5\times 10^{-2}$ & 86.6 & 93.7 & 81.9 & 88.1 & $1.1\times 10^{-3}$ & 0.0155 & 1.07 & 1.19 & 7.94 & 2.28 \\
		\cline{1-11}
		$1\times 10^{-1}$ & 92.1 & 89.1 & 88.0 & 86.1 & $5.0\times 10^{-4}$ & 0.0272 & 1.04 & 1.22 & 5.77 & 2.66 \\
		\cline{1-11}
		$2\times 10^{-2}$ & 81.0 & 78.8 & 79.0 & 78.2 & $2.4\times 10^{-3}$ & 0.009 & 1.04 & 1.17 & 5.08 & 3.15 \\
		\cline{1-11}		
	\end{tabular}\label{tab:OTS}
\end{table*}

\begin{table*}[h!]
	\tiny %\scriptsize
	\renewcommand\arraystretch{1.8}
	\centering	
	\caption{Performance comparison of plain networks and ResNets with $K=1$ in terms of training and testing accuracy, OTS, LSR, and LSS on CIFAR-100.}
	\begin{tabular}{c|c|c|c|c|c|c|c|c|c|c}
		\cline{1-11}
		\multirow{2}*{$\gamma$}&\multicolumn{2}{c|}{Training Accuracy ($\%$)}&\multicolumn{2}{c|}{Testing Accuracy ($\%$)}&\multicolumn{2}{c|}{OTS} &\multicolumn{2}{c|}{LSR}&\multicolumn{2}{c}{LSS}\\
		\cline{2-11}
		~& plain & ResNet & plain & ResNet & plain & ResNet & plain & ResNet & plain & ResNet \\
		\cline{1-11}
		0 & 27.3 & 40.3 & 27.3 & 42.4 & $5.3\times10^{-3}$ & 0.205 & 48.6 & 3.62 & 20.1 & 3.47 \\
		\cline{1-11}
		$1\times 10^{-4}$ & 26.8 & 40.3 & 27.6 & 41.8 & $3.4\times10^{-3}$ & 0.629 & 43.9 & 3.57 & 21.9 & 3.57 \\
		\cline{1-11}
		$5\times 10^{-4}$ & 27.4 & 41.8 & 27.9 & 43.2 & $2.2\times10^{-3}$ & 0.963 & 30.8 & 3.48 & 17.0 & 3.50 \\
		\cline{1-11}
		$1\times 10^{-3}$ & 30.2 & 42.9 & 30.4 & 44.1 & $1.41\times10^{-2}$ & 0.822 & 15.4 & 3.62 & 13.5 & 3.73 \\  
		\cline{1-11}
		$5\times 10^{-3}$ & 53.8 & 55.9 & 50.4 & 55.3 & $6.5\times10^{-3}$ & 0.088 & 1.55 & 2.17 & 7.87 & 2.26 \\
		\cline{1-11}
		$1\times 10^{-2}$ & 68.3 & 67.0 & 58.9 & 59.7 & $3\times10^{-4}$ & $5.70\times10^{-3}$ & 1.13 & 1.52 & 6.99 & 2.01 \\
		\cline{1-11}
		$5\times 10^{-2}$ & 62.4 & 55.0 & 55.1 & 52.1 & $2.4\times10^{-3}$ & $7\times10^{-4}$ & 1.03 & 1.51 & 6.07 & 2.13 \\
		\cline{1-11}
		$1\times 10^{-1}$ & 54.2 & 44.8 & 50.1 & 44.5 & $5\times10^{-4}$ & $4.6\times10^{-3}$ & 1.02 & 1.14 & 6.20 & 2.45 \\
		\cline{1-11}
		$2\times 10^{-2}$ & 12.6 & 14.0 & 11.8 & 12.0 & $2.2\times10^{-3}$ & $2.3\times10^{-3}$ & 1.02 & 1.29 & 5.85 & 3.37 \\
		\cline{1-11}		
	\end{tabular}
\end{table*}

\clearpage

\begin{table*}[t!] 
    \tiny
%	\scriptsize
	\renewcommand\arraystretch{1.8}
	\centering	
	\caption{Performance comparison of plain networks and ResNets with $K=4$ in terms of testing accuracy, OTS, LSR, and LSS on CIFAR-10.}
	\begin{tabular}{c|c|c|c|c|c|c|c|c}
		\cline{1-9}
		\multirow{2}*{$\gamma$}&\multicolumn{2}{c|}{Testing Accuracy ($\%$)}&\multicolumn{2}{c|}{OTS (std)} &\multicolumn{2}{c|}{LSR (std)}&\multicolumn{2}{c}{LSS (std)}\\
		\cline{2-9}
		~& plain & ResNet & plain & ResNet & plain & ResNet & plain & ResNet\\
		\cline{1-9}
		0 & 89.3 & 94.1 & $5.5\times10^{-3}$ ($6.7\times 10^{-3}$) & 1 (0) & 3.40 (0.27) & 1.30 (0.022) & 4.66 (0.33)  & 1.40 (0.02)   \\
		\cline{1-9}
		$5\times 10^{-4}$ & 92.6 & 95.2 & $1.75\times10^{-3}$ ($1.3\times 10^{-3}$) & 1 (0) & 3.33 (0.94) & 1.17 (0.071) & 3.87 (0.62)   & 1.21 (0.08)   \\
		\cline{1-9}
		$1\times 10^{-3}$ & 94.0 & 95.4 & 0 (0) & 0.813 (0.32) & 2.55 (0.97) & 1.16 (0.080) & 3.62 (0.65)   & 1.18 (0.09)    \\  
		\cline{1-9}
		$5\times 10^{-3}$ & 92.2 & 95.5 & $1.25\times10^{-3}$ ($1.1\times10^{-3}$) & 0.760 (0.42) & 2.13 (0.61) & 1.49 (0.34) & 3.36 (0.23)  & 1.70 (0.24)   \\
		\cline{1-9}
		$1\times 10^{-2}$ & 91.2 & 95.0 & 0.0188 (0.027) & 0.950 (0.088) & 1.88 (0.55) & 1.56 (0.32) & 3.63 (0.71)  & 1.87 (0.17)   \\
		\cline{1-9}		
	\end{tabular}
\end{table*}

\begin{table*}[h!]
	\tiny   %\scriptsize
	\renewcommand\arraystretch{1.8}
	\centering	
	\caption{Performance comparison of plain networks and ResNets with $K=4$ in terms of testing accuracy, OTS, LSR, and LSS on CIFAR-100.}
	\begin{tabular}{c|c|c|c|c|c|c|c|c}
		\cline{1-9}		\multirow{2}*{$\gamma$}&\multicolumn{2}{c|}{Testing Accuracy ($\%$)}&\multicolumn{2}{c|}{OTS (std)} &\multicolumn{2}{c|}{LSR (std)}&\multicolumn{2}{c}{LSS (std)}\\
		\cline{2-9}
		~& plain & ResNet & plain & ResNet & plain & ResNet & plain & ResNet \\
		\cline{1-9}
		0 & 57.4 & 74.5 & 0.148 (0.20) & 0.999 ($6.0\times 10^{-4}$) & 3.74 (0.37) & 1.35 (0.09) & 4.60 (0.12)  & 1.38 (0.07)   \\
		\cline{1-9}
		$5\times 10^{-4}$ & 64.2 & 76.4 & 0.005 (0.006) & 1 (0) & 3.56 (0.33) & 1.20 (0.027) & 4.13 (0.61)  & 1.28 (0.04)   \\
		\cline{1-9}
		$1\times 10^{-3}$ & 69.6 & 77.1 & 0.223 (0.39) & 0.995 ($6.4\times 10^{-3}$) & 3.08 (0.44) & 1.21 (0.055) & 3.98 (0.51)  & 1.30 (0.06)  \\  
		\cline{1-9}
		$5\times 10^{-3}$ & 67.8 & 79.7 & 0.057 (0.08) & 0.964 ($6.3\times 10^{-2}$) & 2.79 (0.70) & 1.43 (0.23) & 3.54 (0.74)  & 1.55 (0.18)  \\
		\cline{1-9}
		$1\times 10^{-2}$ & 44.1 & 78.1 & 0.003 (0.001) & 0.859 (0.24) & 2.99 (1.18) & 1.66 (0.36) & 3.87 (0.45)  & 1.82 (0.20)  \\
		\cline{1-9}		
	\end{tabular}
\end{table*}
\clearpage

% that's all folks
\end{document}